\documentclass{article} % For LaTeX2e
\usepackage{iclr2017_workshop,times}
\usepackage{hyperref}
\hypersetup{%
  colorlinks=true,% hyperlinks will be coloured
  linkcolor=green,% hyperlink text will be green
  linkbordercolor=green,% hyperlink border will be red
}
\usepackage{url}
\usepackage{paralist}
\usepackage{graphicx}
\usepackage{subfigure}
\usepackage{times}
\usepackage{wrapfig}
\usepackage{amsmath,amssymb}
\usepackage{caption}
\usepackage[noend]{algpseudocode}
\usepackage{algorithm}
\usepackage[export]{adjustbox}
\usepackage{pbox}
\usepackage[update,prepend]{epstopdf} 
\newcommand\NoDo{\renewcommand{\algorithmicdo}{}}
 \newcommand\NoThen{\renewcommand{\algorithmicthen}{}}

\DeclareMathOperator*{\argmax}{\arg\!\max}
\title{Robustness to Adversarial Examples through an Ensemble of Specialists}
\graphicspath{{./}{figures/}}

\author{Mahdieh Abbasi \& Christian Gagn\'e\\
Computer Vision and Systems Laboratory, Electrical and Computer Engineering Department\\
Universit\'e Laval, Qu\'ebec (Qu\'ebec), Canada \\
\texttt{mahdieh.abbasi.1@ulaval.ca,} \texttt{christian.gagne@gel.ulaval.ca}
}

\begin{document}

\maketitle

\section{Introduction}

Due to the recent breakthroughs achieved by Convolutional Neural Networks (CNNs) for various computer vision tasks~\citep{he2015deep,taigman2014deepface,karpathy2014large}, CNNs are highly regarded technology for inclusion into real-life vision applications. However, CNNs have a high risk of failing due to adversarial examples, which fool them consistently with the addition of small perturbations to natural images, undetectable by the human eyes. 

To mitigate this risk, it has been proposed to train CNNs on both clean training samples and corresponding adversarial examples, generated by some existing algorithms~\citep{goodfellow2014explaining,szegedy2013intriguing,moosavi2015deepfool,sabour2015adversarial}. Although such trained CNNs are robust to specific types of adversaries, they are not necessarily protected from all possible types. To increase the robustness of CNNs, it has been proposed to train them on a \emph{diverse} set of adversaries, generating adversarial examples for any single images with various algorithms~\citep{rozsa2016adversarial}. However, it is still possible to produce other types of adversaries, uncovered by the current set, impacting significantly the reliability of CNNs. Moreover, training on some type of adversaries has been demonstrating to harm the performance on clean test samples~\citep{jin2015robust,moosavi2015deepfool}. 
 
We are rather considering recognition of adversarial examples as an open set recognition problem, where \emph{unknown} samples should be detected and rejected by the underlying models. \citet{bendale2016towards} have adapted CNNs by adding an extra layer designed to recognize the unknown samples, which can be either from unknown classes or fooling adversarial instances from~\citet{nguyen2015deep}. However, as mentioned by the authors, the method fails to detect hard adversaries where the target class and the true class of an adversary are close together, like those generated by Fast Gradient Sign (FGS)~\citep{goodfellow2014explaining} and DeepFool (DF)~\citep{moosavi2015deepfool}. 

We are proposing to use an ensemble of diverse specialists, where speciality is defined according to the confusion matrix. Indeed, we observed that for adversarial instances originating from a given class, labeling tend to be done into a small subset of (incorrect) classes. Therefore, we argue that an ensemble of specialists should be better able to identify and reject fooling instances, with a high entropy (i.e., disagreement) over the decisions in the presence of adversaries. Experimental results obtained confirm this interpretation that a rejection mechanism can provide a means of rendering the system more robust to adversarial examples, rather than trying to classify them properly at any cost.

\section{Specialists+1 Ensemble}

\begin{figure}
\centering
\subfigure[MNIST Confusion Matrix]{\includegraphics[width = 0.495\textwidth, trim=0.4cm 0cm 1cm 1cm, clip=true]{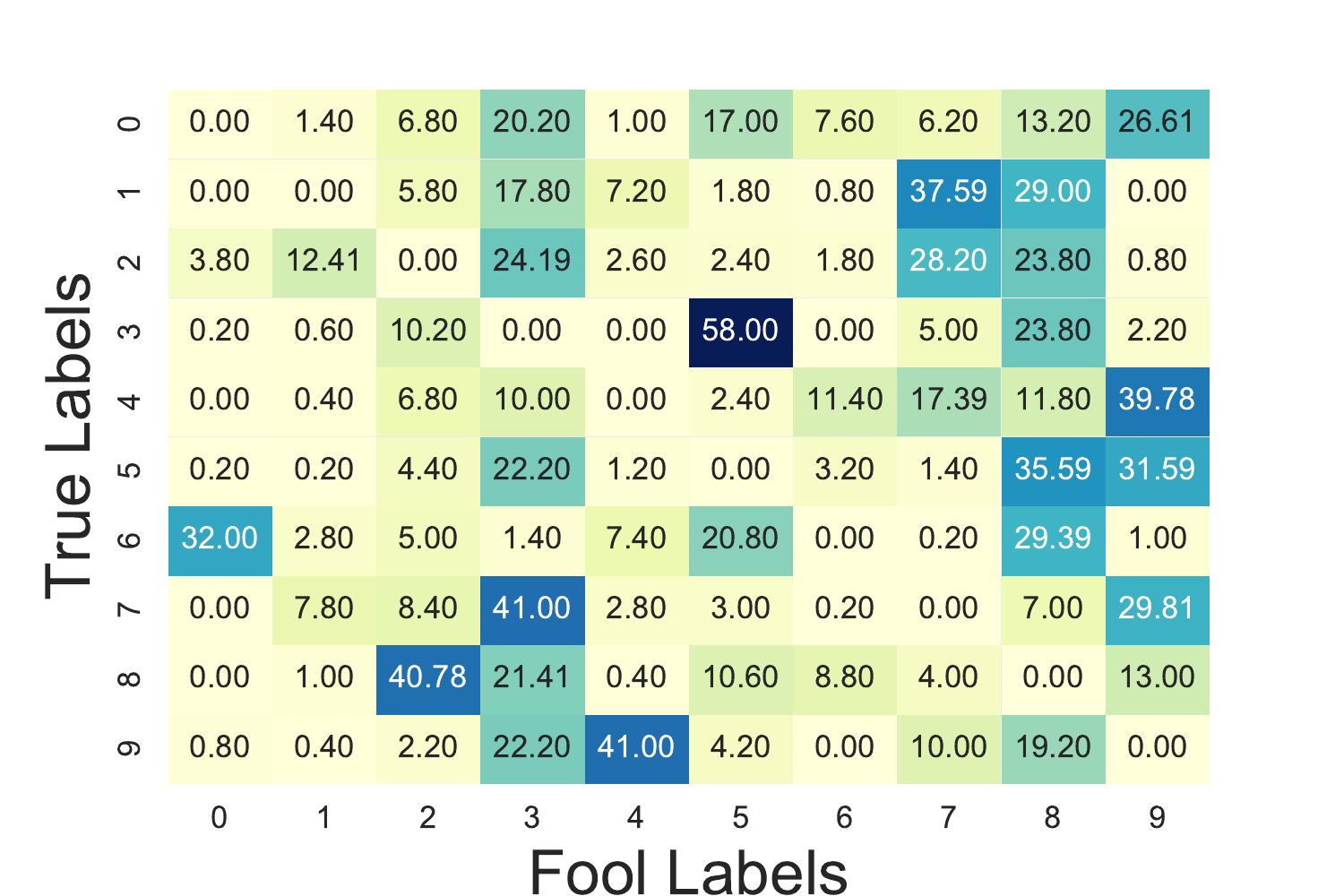}}
\subfigure[CIFAR-10 Confusion Matrix]{\includegraphics[width = 0.495\textwidth,  trim=0.3cm 0cm 1cm 1cm, clip=true]{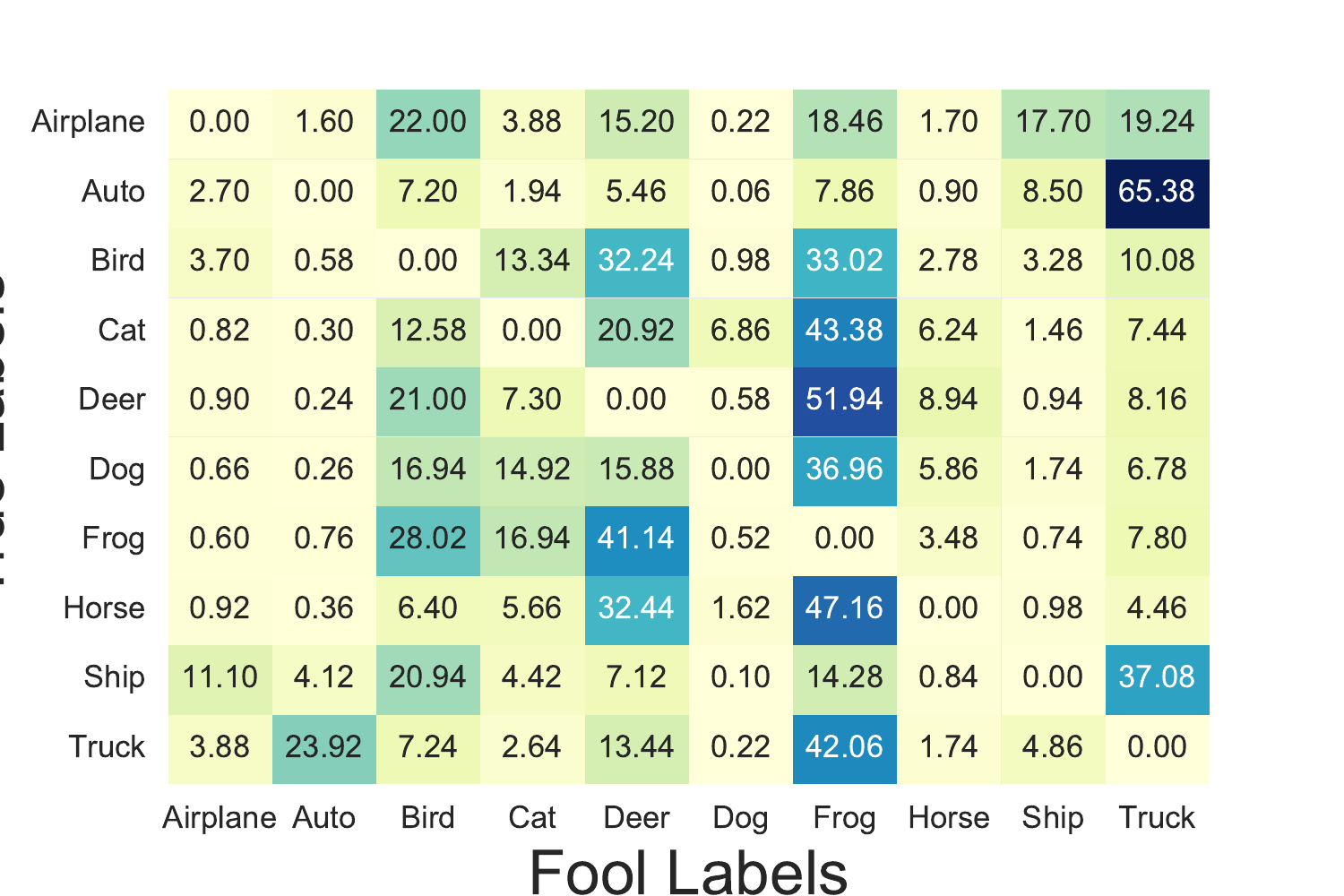}}
%\squeezeup{-0.5cm}
\caption{Confusion matrices of adversaries for (a) MNIST and (b) CIFAR-10. These matrices have been computed from 5000 randomly selected FGS training adversaries (500 per class).}
\label{HistAdver}
\end{figure}

\paragraph{Ensemble construction} The confusion matrices of FGS adversaries (Fig.~\ref{HistAdver}) reveals that samples from each class have a high tendency of being fooled toward a limited number of classes.
From these confusion matrices of training adversaries we define subsets of classes, similarly to \citet{hinton2015distilling} on training clean samples. Considering a classification problem of $K$ classes ($C=\{c_1,c_2,\ldots,c_K\}$), each row of the confusion matrix is used to identify two subsets of classes: 1) the confusing target subset for class $c_i$ (subset $U_i$), which is built by adding classes sequentially in decreasing $c_i$-related confusion values order until at least 80\,\% of confusions are covered, and 2) the remaining classes with lower confusion, formed as subset $U_{i+K}=C\setminus U_i$. Duplicate subsets should be ignored, although we encountered none of them for MNIST and CIFAR-10.

For each of these class subsets, a specialist CNN is trained on samples from the associated classes, instances from the other classes being ignored. The ensemble also includes a generalist CNN trained on the complete labels (subset $U_{2K+1}$), hence the name ``specialists+1 ensemble''. 

\paragraph{Voting mechanism}
Using the generalist to activate the related specialists is not possible as it is usually being fooled by adversaries. In Algorithm~\ref{Votingmechanism}, we propose a voting mechanism to compute the final prediction. 
\begin{algorithm}[tb]
\caption{Voting Mechanism}
\label{Votingmechanism}
\begin{algorithmic}[1]
\Require Ensemble $\mathcal{H}=\{h^{1},\dots,h^{M}\}$ with $h^{j}\in \mathbb{R}^{K}$, label subsets $\mathcal{U}=\{U_1,\ldots,U_M\}$, input $\mathbf{x}$
\Ensure Final prediction $\bar{h}(\mathbf{x})\in\mathbb{R}^{K}$
\NoDo
\NoThen
\State $v_k(\mathbf{x}) \gets \sum_{j=1}^{M} \mathbb{I}[c_k = \argmax_{i=1}^K h^j_i(\mathbf{x})],~k=1,\dots,K\label{equation_alg}$
\State $k^* \gets \argmax_{k=1}^K v_k(\mathbf{x})$
\If $v_{k^*}(\mathbf{x}) = K+1$
  \State $\mathcal{S} \gets \{h^i\in\mathcal{H}~|~c_{k^*}\in U_i\}$
  \State $\bar{h}(\mathbf{x}) \gets \frac{1}{K+1}\sum_{h^i\in\mathcal{S}} h^i(\mathbf{x})$
\Else
  %\State $\mathcal{S} \gets \{h^i\in\mathcal{H}~|~c_k\in U_i\}$
  \State $\bar{h}(\mathbf{x}) \gets \frac{1}{M}\sum_{h^i\in\mathcal{H}} h^i(\mathbf{x})$
\EndIf\\
\Return $\bar{h}(\mathbf{x})$
\end{algorithmic}
\end{algorithm}
As each class $c_{i}$ appears $K+1$ times in $M$ subsets of classes, \emph{the maximum expected number of votes} to class $c_{i}$ is $K+1$. Also, we define $v_{i}(\mathbf{x})$ as \emph{the actual number of votes} to class $c_{i}$ for a given input image $\mathbf{x}$ (the equation in line~\ref{equation_alg} of the algorithm~\ref{Votingmechanism}). If only one class has its actual number of votes equal to its maximum expected number of votes, i.e., $v_i(\mathbf{x})=K+1$, it means that all $K$ related specialists and the generalist agree to vote to the winner class. Then only those CNNs voting for the winner class should be activated in order to compute the final prediction. Otherwise, if none of the classes obtain their maximum expected number of votes, it means that at least one of the individuals was fooled. So, some votes are incorrectly distributed between different classes. In the presence of such entropy, where there is no winner class, all of the individuals should be activated to compute the final prediction.

\section{Empirical Evaluation}

\paragraph{Networks and datasets} Similarly to~\citet{hinton2012improving}, we used CNNs with three convolutional layers having 32, 32, and 64 filters respectively, and a fully-connected layer followed by a softmax. The networks are trained on usual training and testing sets of MNIST and CIFAR-10, without any data augmentation. See the Appendix section for full details on the network architecture and hyper-parameters used for each dataset. Note that all CNNs presented in the experiments have an identical architecture.

\paragraph{Experiments} We compared our proposed specialists+1 ensemble with a pure ensemble, which consists of 5 generalist CNNs with different random initializations, and a naive CNN*, whose weights initialization is different from GA-CNN (i.e., the CNN used to generate the adversaries). Using this GA-CNN, three types of adversaries, namely FGS, DF, and \citet{szegedy2013intriguing} adversaries, are generated for correctly classified clean test samples.

\begin{figure}

\subfigure[Naive CNN*]{\label{MNIST:Naive1}\includegraphics[width=0.3\textwidth, trim=2cm 0cm 2cm 1cm, clip=true]{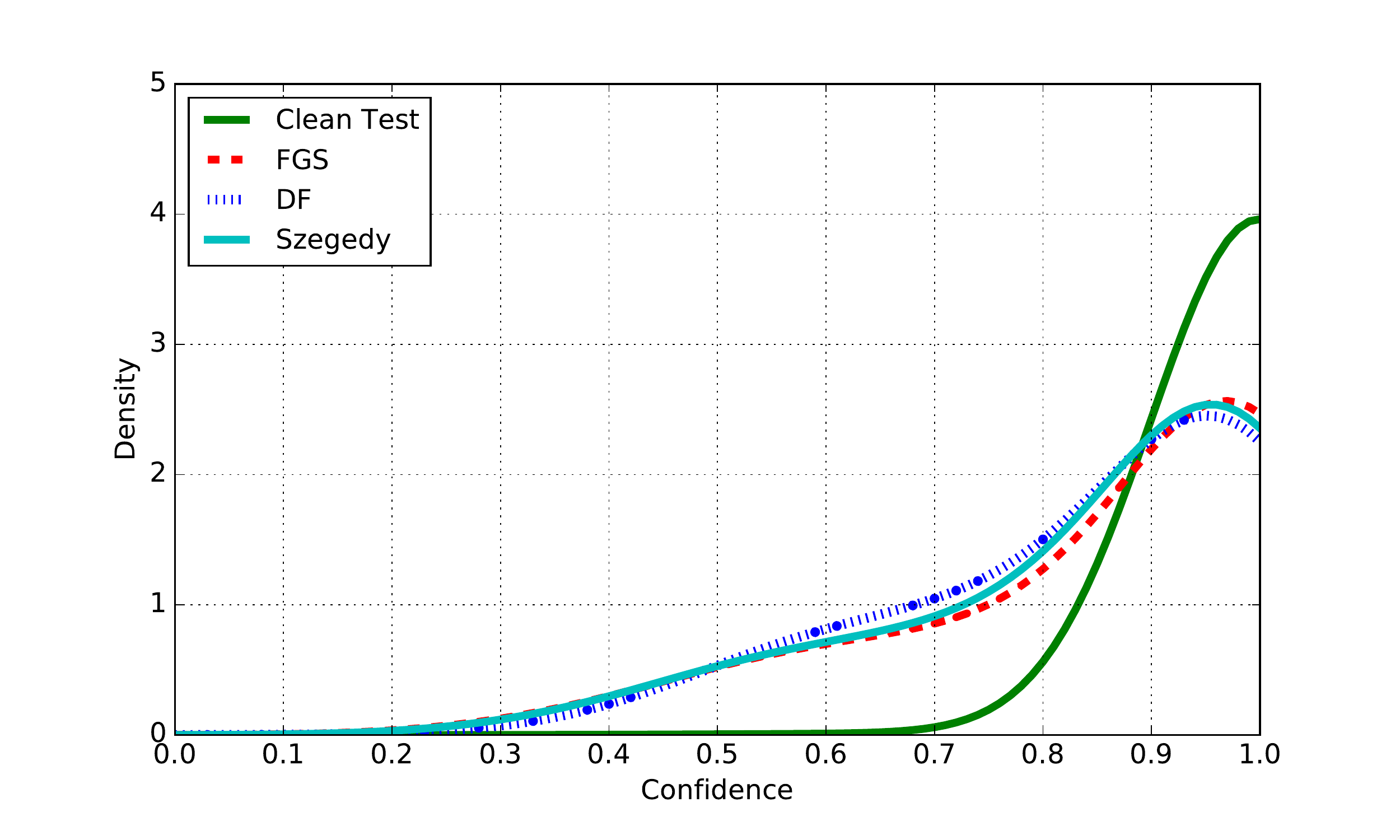}} \hfil
\subfigure[Pure ensemble]{\label{MNIST:pure1}
\includegraphics[width=0.3\textwidth, trim=2cm 0cm 2cm 1cm, clip=true]{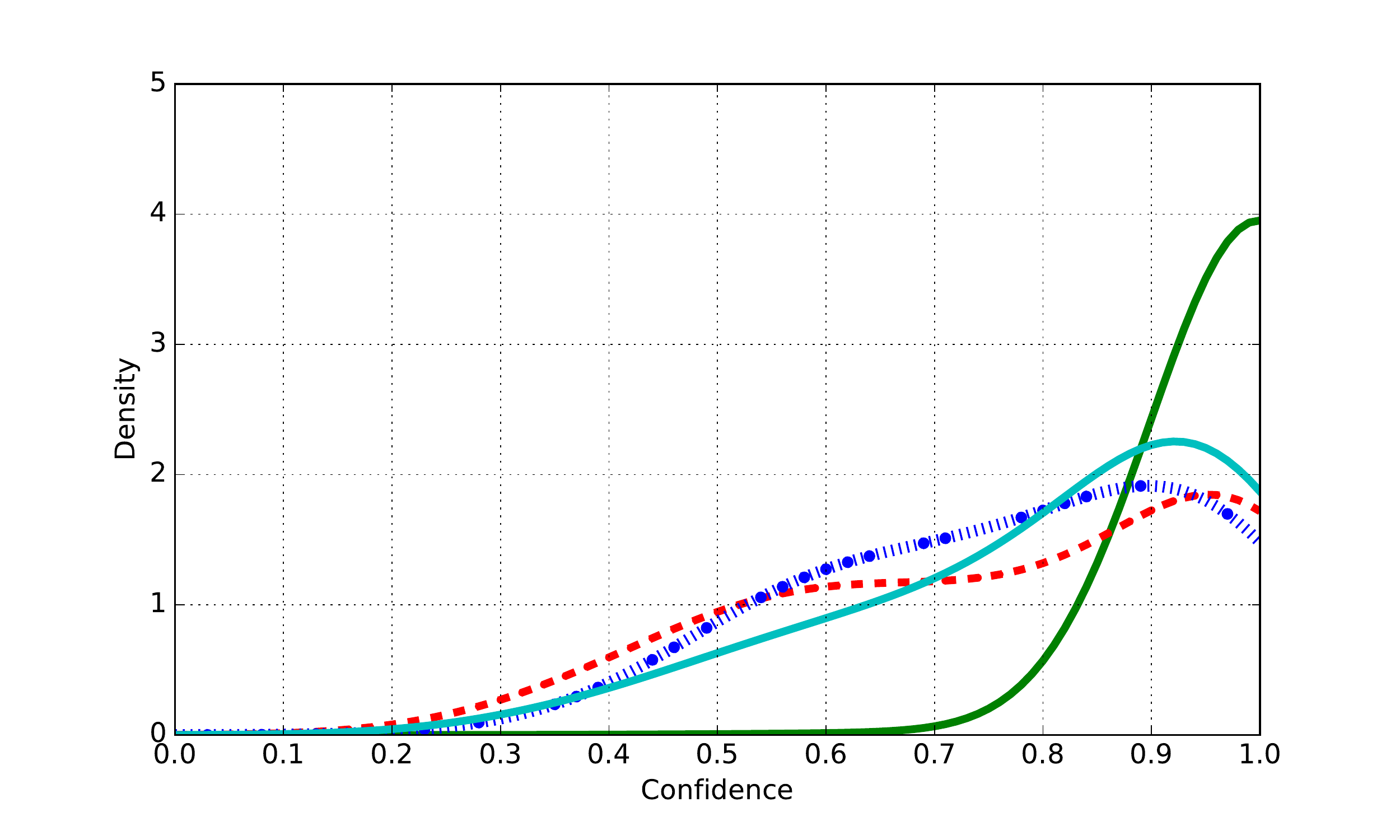}}\hfil
\subfigure[Specialists+1]{
\label{MNIST:Our1}
\includegraphics[width=0.3\textwidth,trim=2cm 0cm 2cm 1cm, clip=true]{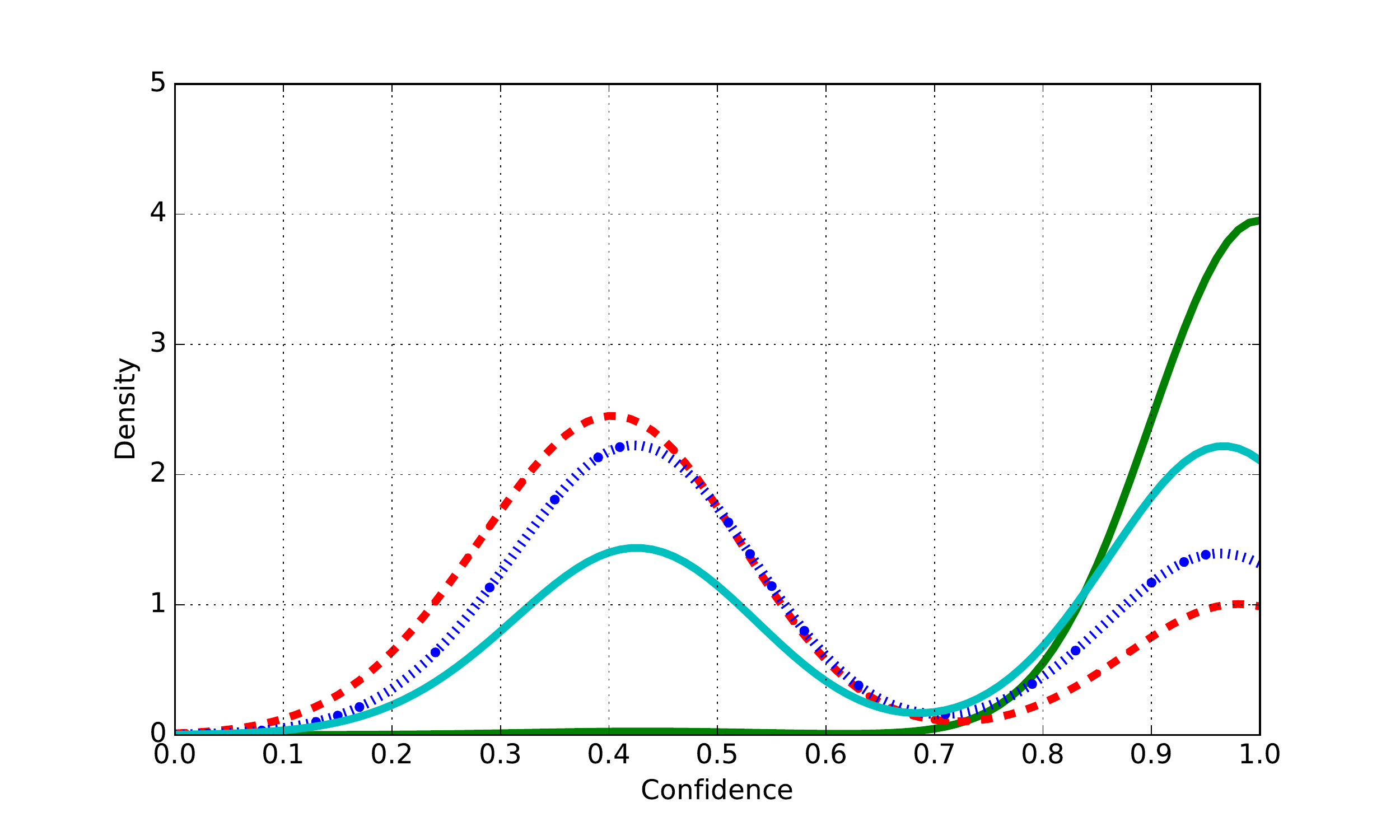}}

\caption{Confidence densities on MNIST: (a) naive CNN*, (b) pure ensemble, and (c) specialists+1.}
\label{MNIST-DEN}
\end{figure}

\begin{figure}
\subfigure[Naive CNN*]{\label{Cifar-Naive1}
\includegraphics[width=0.3\textwidth, trim=2cm 0cm 2cm 1cm, clip=true]{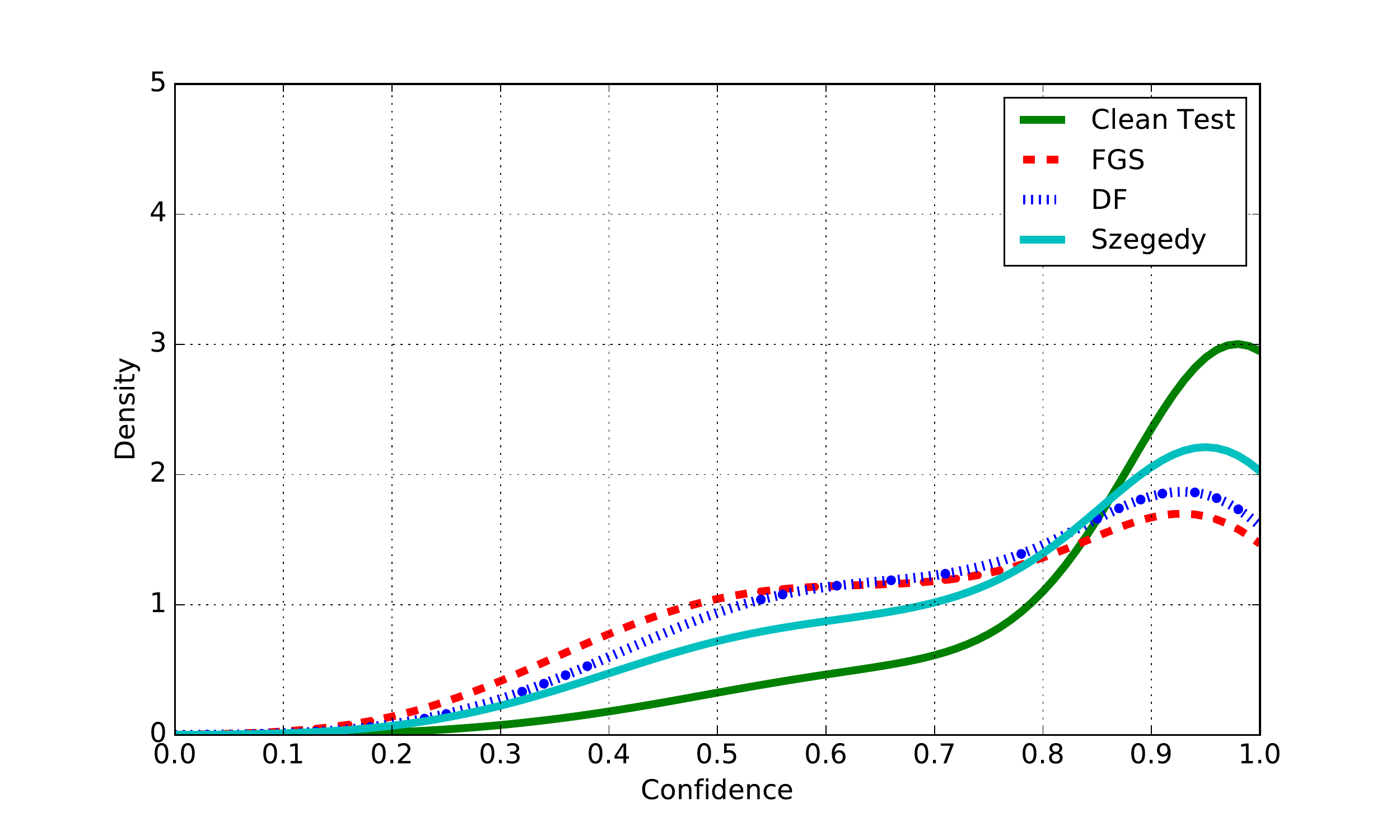}}\hfil
\subfigure[Pure ensemble]{\label{Cifar-pure1}
\includegraphics[width=0.3\textwidth, trim=2cm 0cm 2cm 1cm, clip=true]{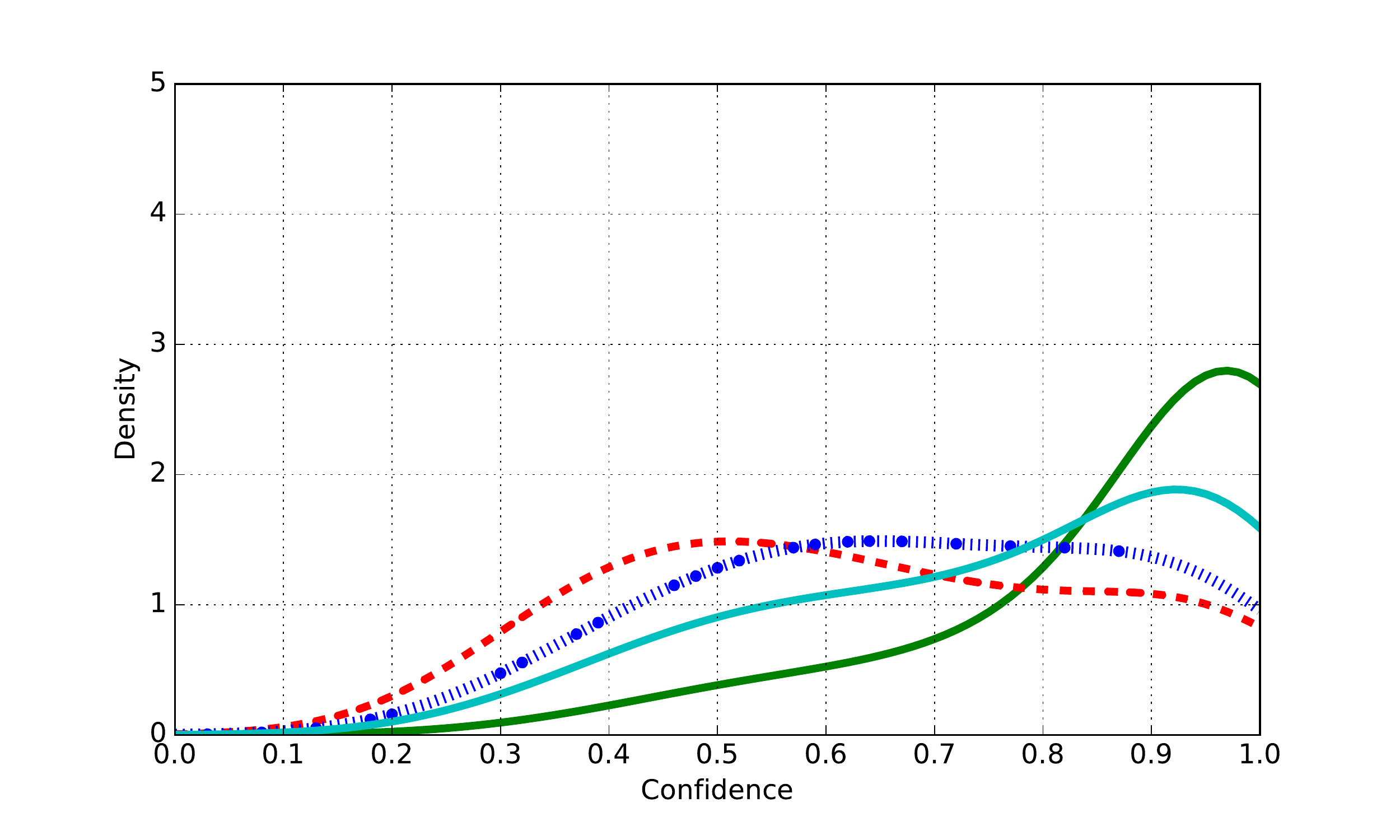}}\hfil
\subfigure[Specialists+1]{\label{Cifar-Our1}
\includegraphics[width=0.3\textwidth, trim=2cm 0cm 2cm 1cm, clip=true]{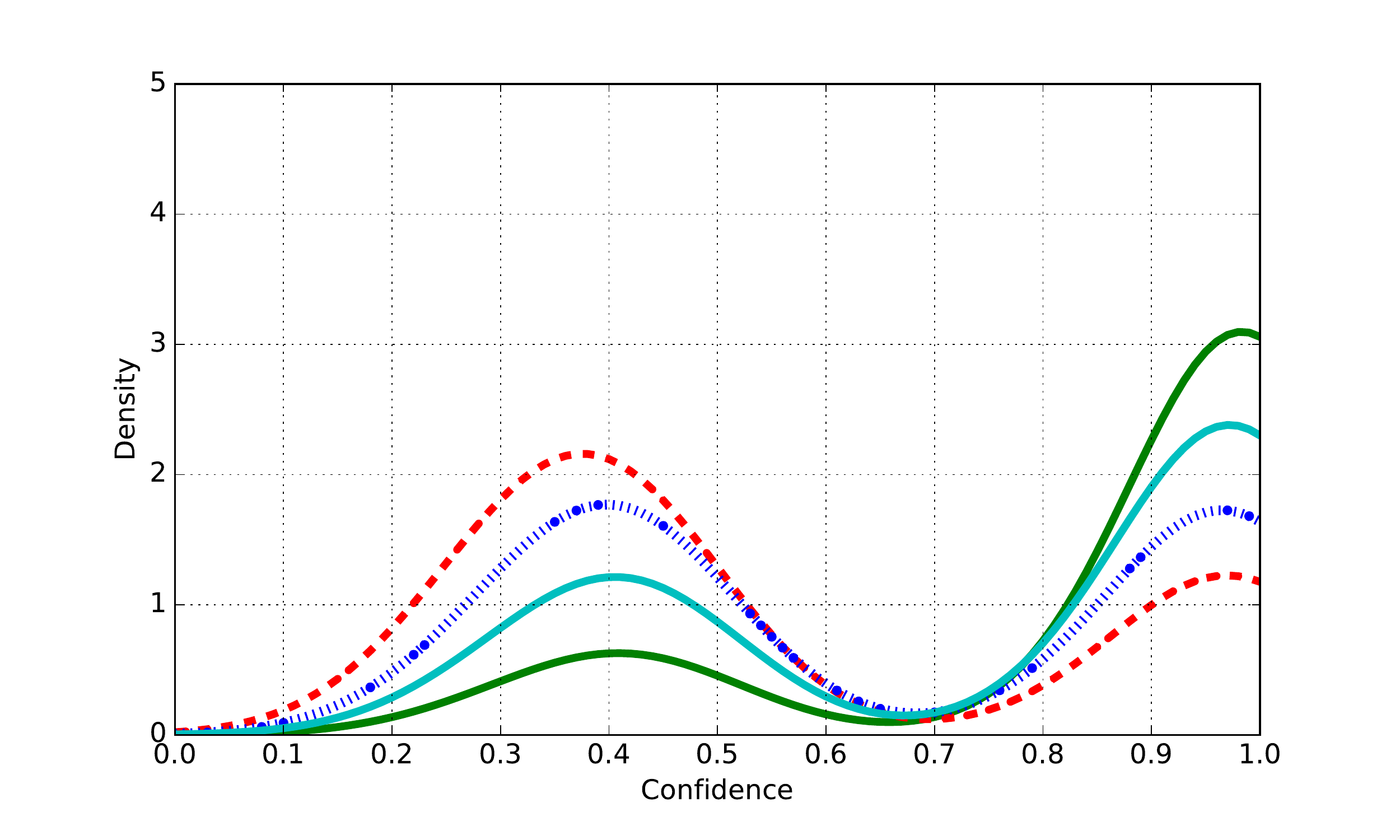}}
\caption{Confidence densities on CIFAR-10: (a) naive CNN*, (b) pure ensemble, and (c) specialists+1.}
\label{Cifar-DEN}
\end{figure}

Naive CNN*, pure ensemble, and specialists+1 ensemble are compared according to their distributions of confidence on correctly classified clean test samples and their corresponding adversaries for MNIST and CIFAR-10 in Fig.~\ref{MNIST-DEN} and \ref{Cifar-DEN}, respectively. According to these observations for MNIST, specialists+1 successfully provides significantly lower confidence for most of the misclassified adversaries, regardless of their types, than naive CNN* and the pure ensemble. Also, as it can be seen from Fig.~\ref{MNIST:Naive1}, \ref{MNIST:pure1}, and \ref{MNIST:Our1} that the distribution of confidence on MNIST correctly classified clean test samples by these three frameworks are roughly similar. For CIFAR-10, we observed the same behavior as MNIST for adversaries (Fig.~\ref{Cifar-DEN}). But for correctly classified clean test samples, specialists+1 shifts some of these samples to lower confidence (the green curve in Fig.~\ref{Cifar-Our1}).

Although the specialists+1 is not trained from any adversaries, it appears able to automatically reduce the confidence of predictions for most of the misclassified adversaries, regardless of their types, while preserving up to some point the confidence on clean samples. However, it reduces the confidence of a few clean test samples. Therefore, developing a learning model that is not confident about unknown samples but yet is confident about known samples can be a used to identify and reject adversaries. We depicted the effect of rejecting low confidence adversaries on the error rates of different types of adversaries in Fig.~\ref{ErrExprmnt-woGAN}, in Appendix due to space consideration.

%\subfigure[Confidence densities on MNIST]{\includegraphics[width=0.4\textwidth]{MNIST_Our_caption}}\\
%\subfigure[Confidence densities on CIFAR-10]{\includegraphics[width=0.4\textwidth]{cifar_Our_wo}}
%\caption{Confidence densities of specialists+1 ensemble on MNIST and CIFAR-10 over clean test samples and three adversarial.}

\paragraph{Conclusion} In brief, without training from adversaries and by leveraging diversity in ensembles by a specialization over the labels, the specialists+1 ensemble approach is able to better discriminate between legitimate samples and adversarial instances. The approach is better at rejecting adversaries while accepting clean samples based on confidence, compared to the pure ensemble and Naive CNN*. That is an important matter in order to increase robustness of CNNs to carefully crafted attacks, preferring to refuse processing suspicious instances rather than being fooled by carefully crafted attacks. As future work, we will compare our approach with CNN explicitly trained to being robust to specific types of adversaries.

%\begin{table}[h!]
%\label{adversariesFGS}
%\centering
%\begin{adjustbox}{width=\textwidth,center=\textwidth}
%\begin{tabular}{lcccc|cccc}
%&\multicolumn{4}{c|}{MNIST}&\multicolumn{4}{c}{Cifar-10} \\
%\cline{2-9}
% & \pbox{20cm}{clean Test}& \pbox{20cm}{FGS}& \pbox{20cm}{DF}&\pbox{20cm}{Szegedy}&\pbox{20cm}{clean Test}& \pbox{20cm}{FGS}& \pbox{20cm}{DF}&\pbox{20cm}{Szegedy}\\
%\hline 
%
%Naive CNN*&\textbf{99.76} &86.04&85.03&85.24&\textbf{86.78}&74.19&76.95&81.19\\
%Pure ensemble&99.71&77.19&71.95&67.33&82.74&63.88&65.50&63.25\\
%Specialists+1 (20S+1G)&99.52&\textbf{54.17}&\textbf{49.23}&\textbf{44.06}&80.31&\textbf{55.69}&\textbf{54.59}&\textbf{44.48}\\
%\hline
%\end{tabular}
%\end{adjustbox}
%
%\caption{Average confidences (shown in \%) of correctly classified clean test samples and their corresponding adversaries by different approaches without any threshold. }
%\label{avgConf}
%\squeezeup{-0.3cm}
%\end{table}

\clearpage

\subsubsection*{Acknowledgments}

This work was made possible through funding from NSERC-Canada, MITACS, and E Machine Learning Inc. Computational resources were provided by Compute Canada / Calcul Qu\'ebec and by a GPU grant from NVIDIA. The authors are also grateful to Annette Schwerdtfeger for proofreading this manuscript.

\bibliography{ref}
\bibliographystyle{iclr2017_workshop}

\clearpage

\appendix
\section{Appendix}

\subsection{Experimental Procedures}

We consider a CNN with three convolutional layers and one fully connected layer, where each convolutional layer is interlaced with ReLU, local contrast normalization, and a pooling layer. For regularization, dropout is used at the last layer, i.e., fully-connected layer, with $p=0.5$. All of the hyper parameters such as initial learning rate, training schedule, and so on are set according to~\citet{hinton2012improving}.

\paragraph{MNIST} This dataset contains grayscale images of size 28x28, where each image holds a handwritten digit. The training and test sets have 60,000 and 10,000 samples, respectively. All of the images are scaled to $[0,1]$. 150 epochs for training with batch size 128 and the initial learning rate 0.1 with momentum 0.9 are exploited. The learning rate is decayed by factor 10 twice during training, at epochs 50 and 100.

\paragraph{CIFAR-10} This dataset consists of 50,000 RGB images of size 32x32 as training set and 10,000 32x32 RGB images as test set. Each image contains one object from one of 10 classes. All images, either from train or from test sets, are scaled to $[0,1]$, then normalized by mean subtraction, where the mean is computed over the training set. Like MNIST, 150 epochs with batch size 128 and the initial learning rate is 0.01 with momentum of 0.9. The learning rate decays twice by factor 10 shortly before terminating training, at epochs 120 and 130. 
 
\subsection{Generating Adversaries}

Fast Gradient Sign (FGS)~\citep{goodfellow2014explaining}, DeepFool (DF)~\citep{moosavi2015deepfool}, and the algorithm proposed by~\citep{szegedy2013intriguing} are used for adversarial example generation. The latter algorithm finds minimum required perturbations at a high computational cost, while FGS and DF generates adversaries significantly faster, i.e., less than 3 iterations.

Using the GA-CNN (the baseline CNN for generating adversaries), correctly classified clean samples are identified then used for generating adversarial examples. Therefore, 9943 and 8152 adversaries are generated from MNIST and CIFAR-10 test sets, respectively. The optimal values for hyper parameters of FGS and~\citet{szegedy2013intriguing} are obtained for each dataset such that GA-CNN misclassifies 100\% of the correctly classified clean samples after adding perturbations.

\begin{figure}[b]
\begin{center}
\includegraphics[width=0.5\textwidth]{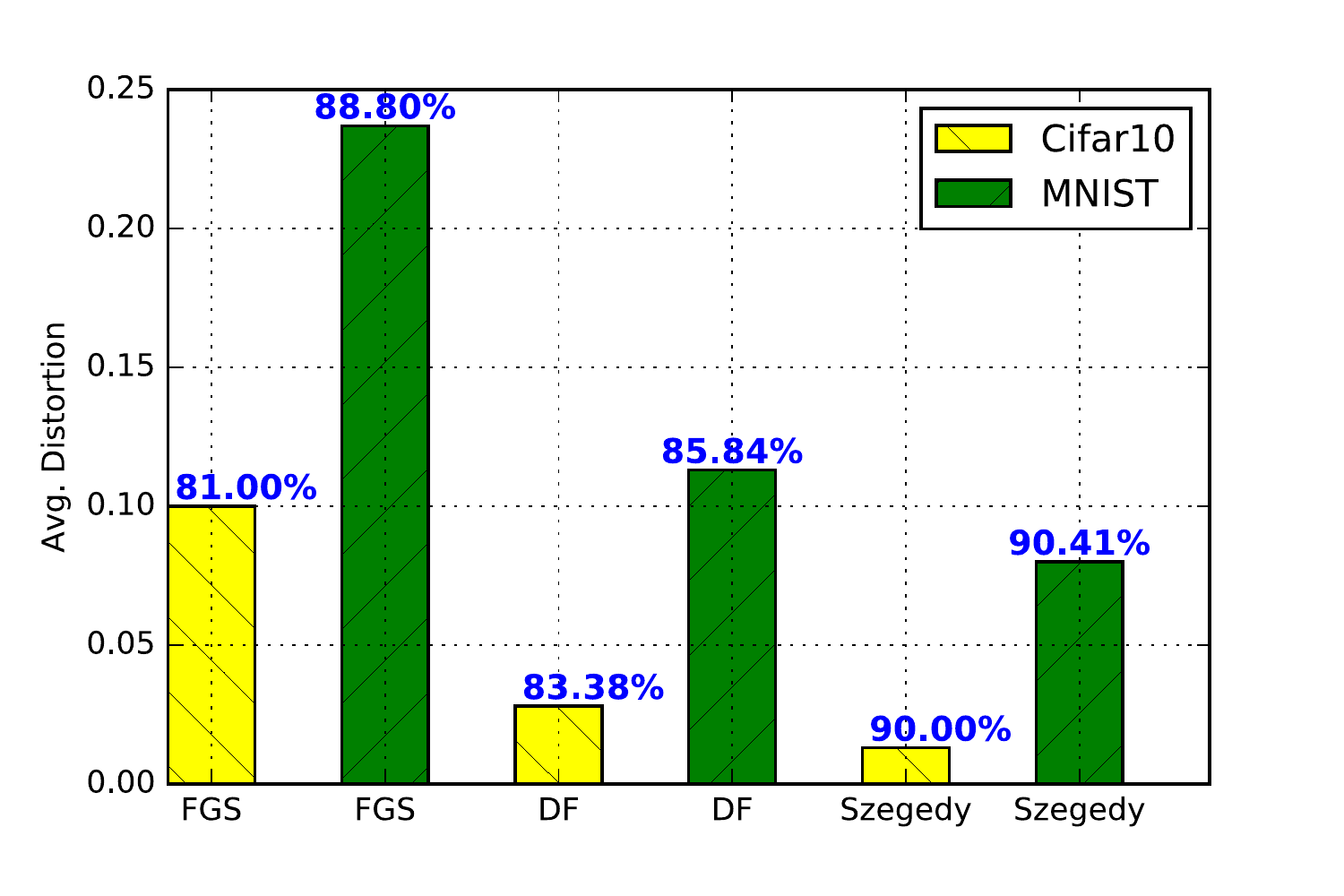}
\caption{Average distortion to MNIST and CIFAR-10 samples by FGS, DF, and~\citet{szegedy2013intriguing}. The average misclassification confidences are shown by blue text.}
\label{adversaries}
\end{center}
\end{figure}

In Fig.~\ref{adversaries}, the average distortions (perturbations) generated by each algorithm for MNIST and CIFAR-10 are depicted. As well, their average misclassification confidences are written in blue. Distortion is measured by $\sqrt{\frac{\sum_{i=1}^D (x_i-x_i')^{2}}{D}}$ for each pair of clean sample ($\mathbf{x}\in \mathbb{R}^{D}$) and its corresponding adversary ($\mathbf{x}'$). 

\subsection{Extra Experimental Results}

Let $h(\mathbf{x})=[h_1(\mathbf{x}),\ldots,h_K(\mathbf{x})]$ be a multi-classification system (e.g., single classifier $h(\mathbf{x})$ or ensemble $\bar{h}(\mathbf{x})$) trained on clean training samples. Like~\citet{bendale2016towards}, we consider a threshold ($\tau$) for rejecting instances with low confidence, assigning them to a reject class $c_{K+1}$. The following classical decision function is used:
\begin{equation}
r(\mathbf{x}) =
\begin{cases}
\argmax\limits_{c_{i}\in C} h_i(\mathbf{x}) & \text{if}~\max\limits_{c_{i}\in C} h_i(\mathbf{x}) \geq\tau\\
c_{K+1}  & \text{otherwise}
\end{cases}.
\end{equation}
Two types of errors should be considered: error $E_{D}$ on the clean set $\mathcal{D}=\{\mathbf{x}_i,y_i\}_{i=1}^N$, and error $E_{A}$ on the adversaries set $\mathcal{A}=\{\mathbf{x}'_i,y'_i\}_{i=1}^{N'}$ ($y'_i$ is the true label of  $\mathbf{x}'_i$). Error $E_{D}$ takes into account both clean samples that are misclassified and correctly classified rejected clean samples:
\begin{equation}
E_{D}=\frac{1}{N}\sum_{i=1}^{N} \left(\mathbb{I}[r(\mathbf{x}_{i})\neq y_i] + \mathbb{I}[r(\mathbf{x}_{i})=c_{K+1} \wedge \argmax\limits_{c_{j}\in C} h_j(\mathbf{x_i})=y_i]\right).\label{ED}
\end{equation}
Error $E_{A}$ considers misclassified adversarial instances that are not rejected:
\begin{equation}
E_{A}= \frac{1}{N'}\sum_{i=1}^{N'} \mathbb{I}[r(\mathbf{x}'_{i})\neq y'_i \wedge r(\mathbf{x}'_{i})\neq c_{K+1}].\label{EA}
\end{equation}
%In Table~\ref{avgConf}, the average confidence of correctly classified clean test samples and their corresponding adversaries are shown for MNIST and Cifar10. The average confidences (shown in~\%) are obtained with a threshold of 0, i.e. no threshold, for further comparison. By proposing specialists+1 ensemble, we attempt to reduce the confidence of adversaries, while preserving the confidence of clean samples. As it can be seen from Table~\ref{avgConf}, specialists+1 ensemble in all types of adversaries significantly reduces the average confidences of the adversaries. However, it reduces the average confidence of clean test samples by 0.24\% and 6.47\% for MNIST and Cifar10 respectively, in comparison to ``Naive CNN*''. 
%\begin{table}[ht!]
%\label{adversariesFGS}
%\centering
%\begin{adjustbox}{width=\textwidth,center=\textwidth}
%\begin{tabular}{lcccc|cccc}
%&\multicolumn{4}{c|}{MNIST}&\multicolumn{4}{c}{Cifar-10} \\
%\cline{2-9}
% & \pbox{20cm}{clean Test}& \pbox{20cm}{FGS}& \pbox{20cm}{DF}&\pbox{20cm}{Szegedy}&\pbox{20cm}{clean Test}& \pbox{20cm}{FGS}& \pbox{20cm}{DF}&\pbox{20cm}{Szegedy}\\
%\hline 
%
%Naive CNN*&\textbf{99.76} &86.04&85.03&85.24&\textbf{86.78}&74.19&76.95&81.19\\
%Pure ensemble&99.71&77.19&71.95&67.33&82.74&63.88&65.50&63.25\\
%Specialists+1 ensemble&99.52&\textbf{54.17}&\textbf{49.23}&\textbf{44.06}&80.31&\textbf{55.69}&\textbf{54.59}&\textbf{44.48}\\
%\hline
%\end{tabular}
%\end{adjustbox}
%
%\caption{Average confidences (\%) of correctly classified clean test samples and their corresponding adversaries by different approaches. }
%\label{avgConf}
%\end{table}

\begin{figure}
\centering
\subfigure[{$E_{D}(\%)$ on MNIST clean test set}]{\label{Err_MNISTcln}\includegraphics[width=0.49\textwidth]{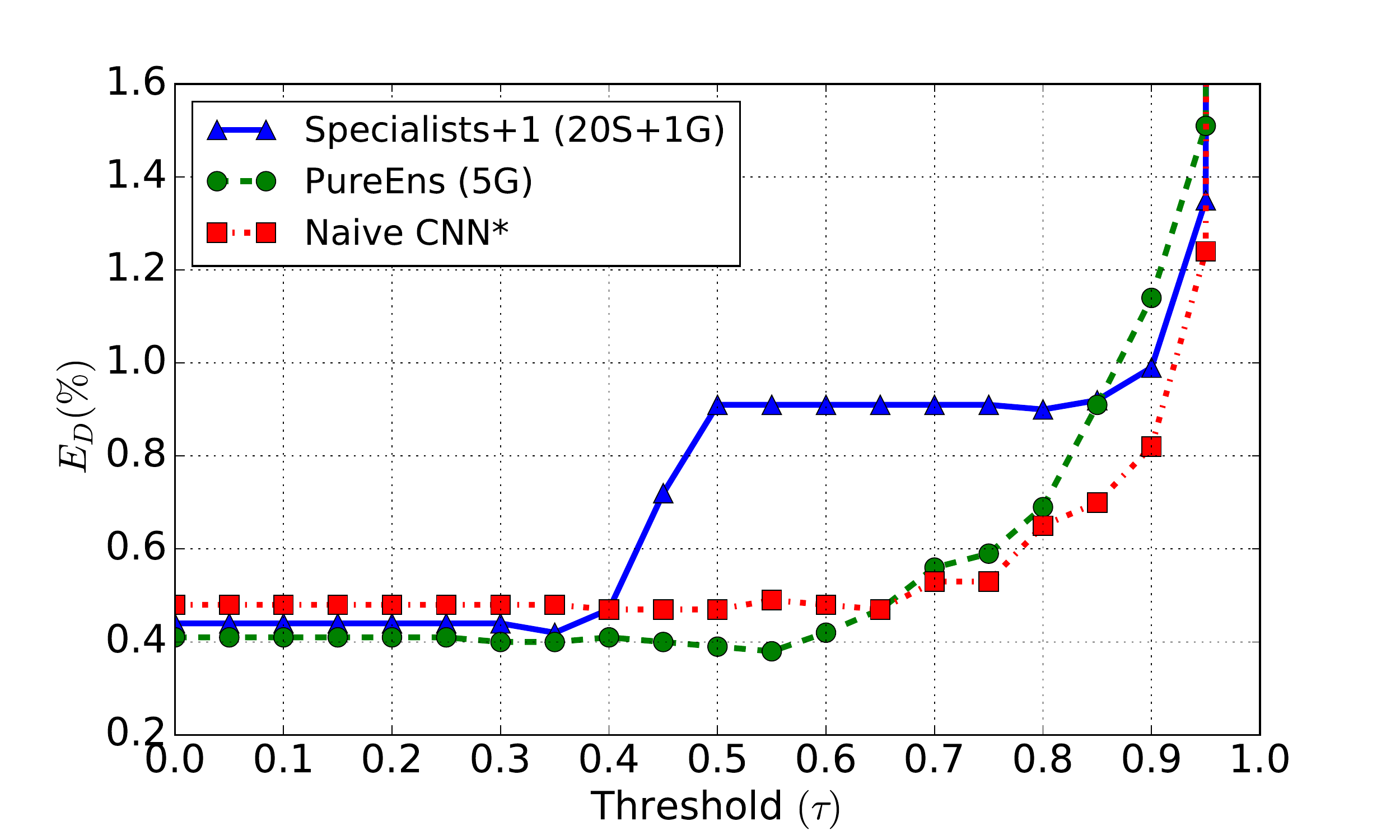}}
\hfil
\subfigure[{$E_{A}(\%)$ on MNIST FGS adversaries}]{\includegraphics[width=0.49\textwidth]{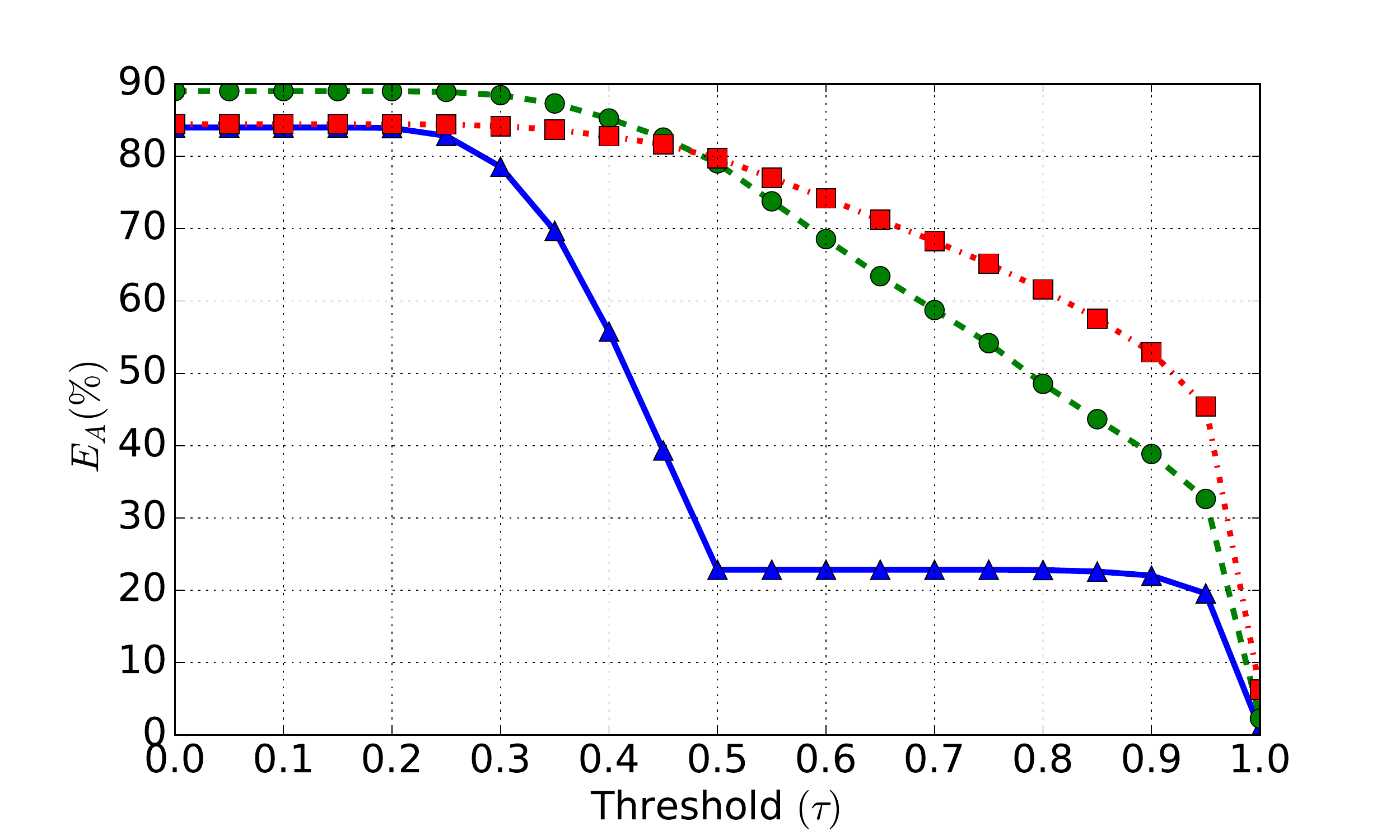}}\\
\subfigure[{$E_{A}(\%)$ on MNIST DF adversaries}] {\includegraphics[width = 0.49\textwidth]{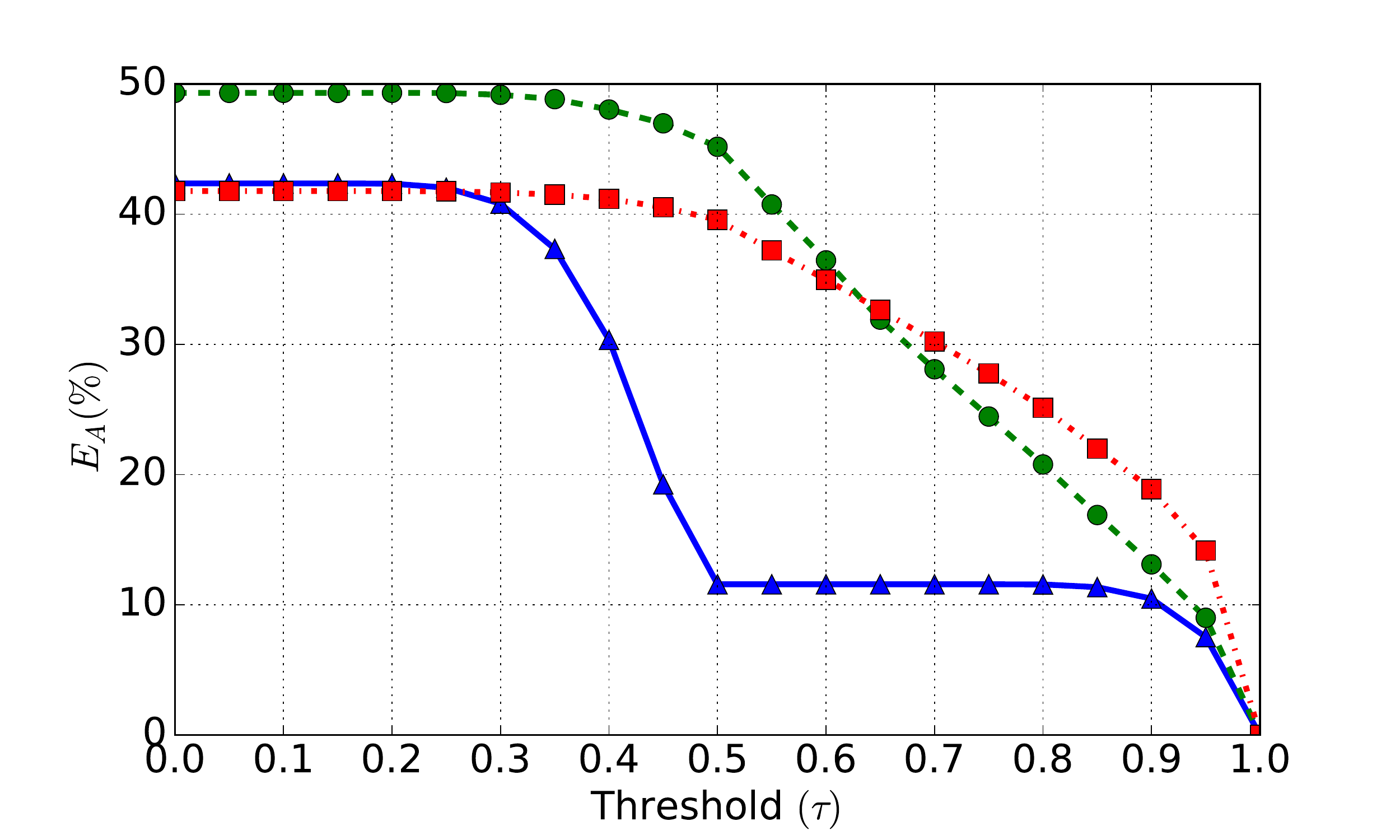}}\hfil
\subfigure[{$E_{A}(\%)$ on MNIST adversaries by~\citet{szegedy2013intriguing}}]{\includegraphics[width = 0.49\textwidth]{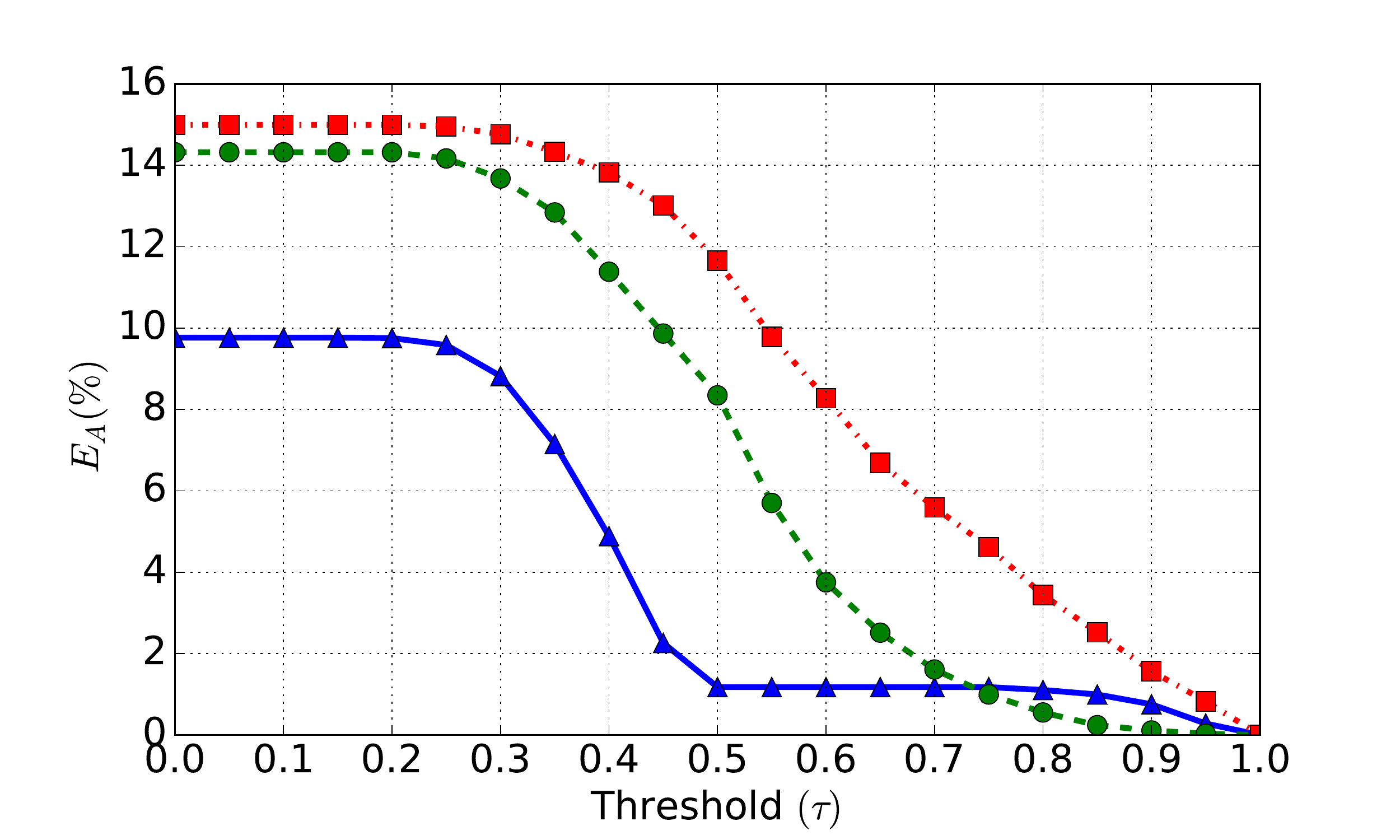}}
\\
\subfigure[{$E_{D}(\%)$ on CIFAR-10 clean samples}]{\label{Err_Cifarcln}\includegraphics[width=0.49\textwidth]{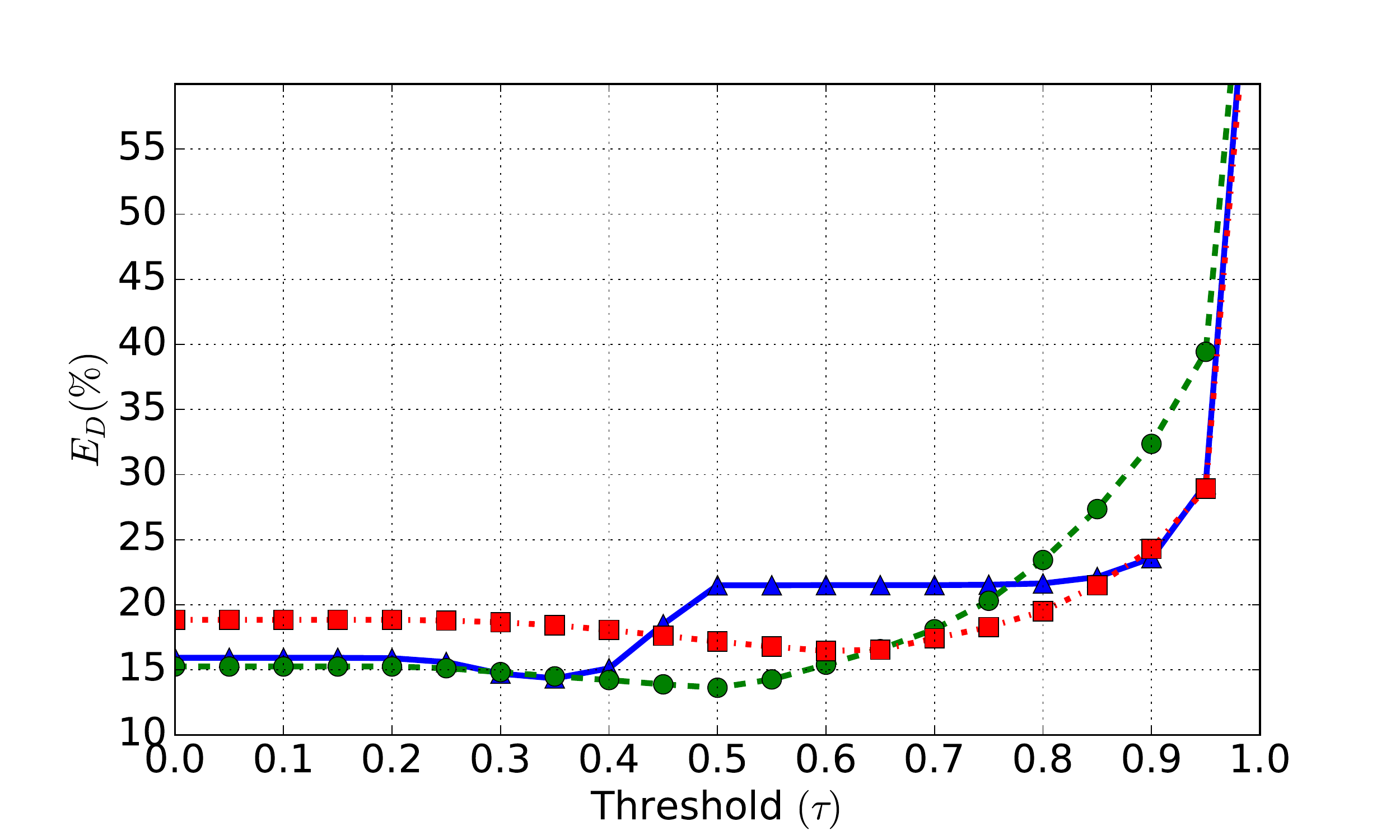}}
\hfil
\subfigure[{$E_{A}(\%)$ on CIFAR-10 FGS adversaries}]{\includegraphics[width = 0.49\textwidth]{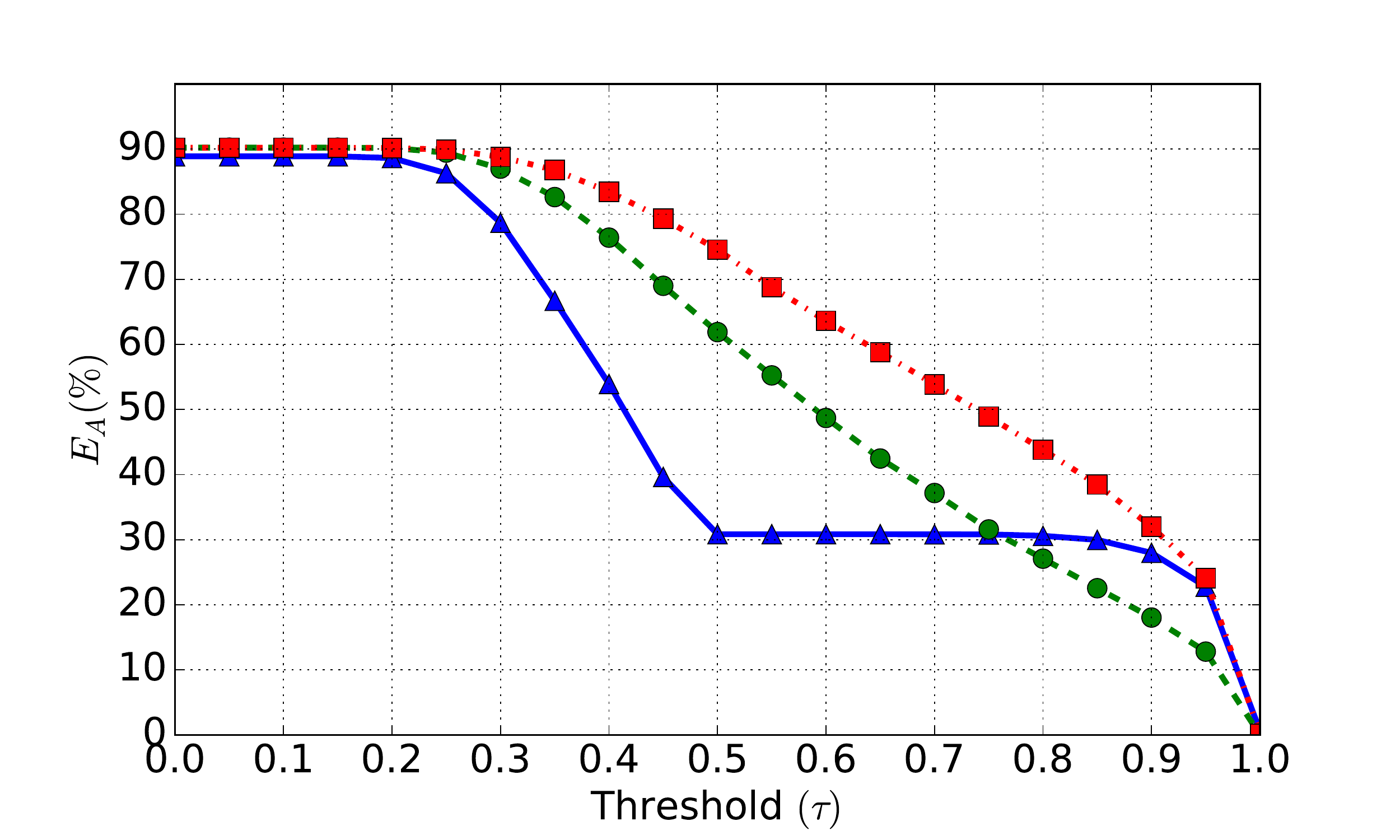}}\\
\subfigure[{$E_{A}(\%)$ on CIFAR-10 DF adversaries}]{\includegraphics[width = 0.49\textwidth]{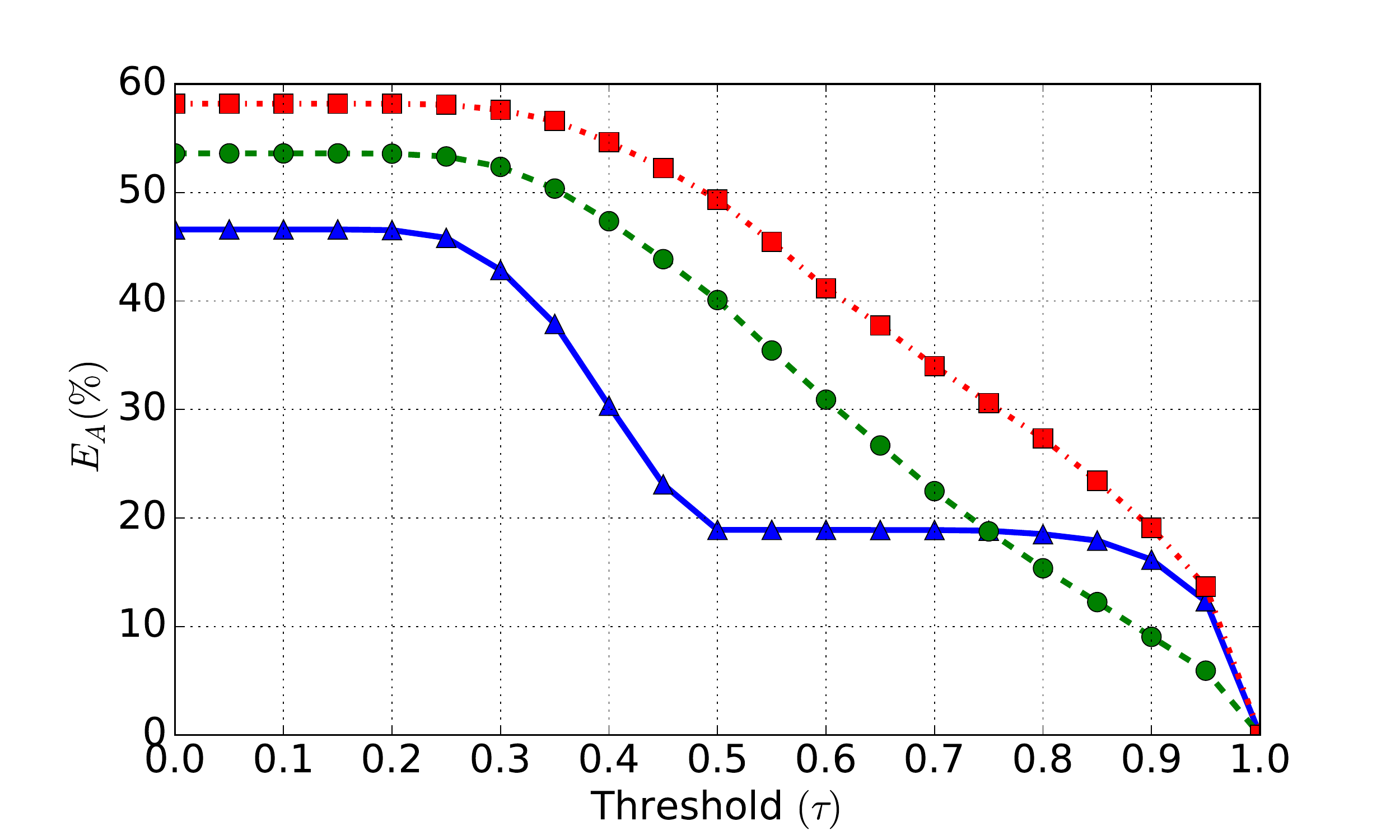}}\hfil
\subfigure[{$E_{A}(\%)$ on CIFAR-10 adversaries by~\citet{szegedy2013intriguing}}]{\includegraphics[width = 0.49\textwidth]{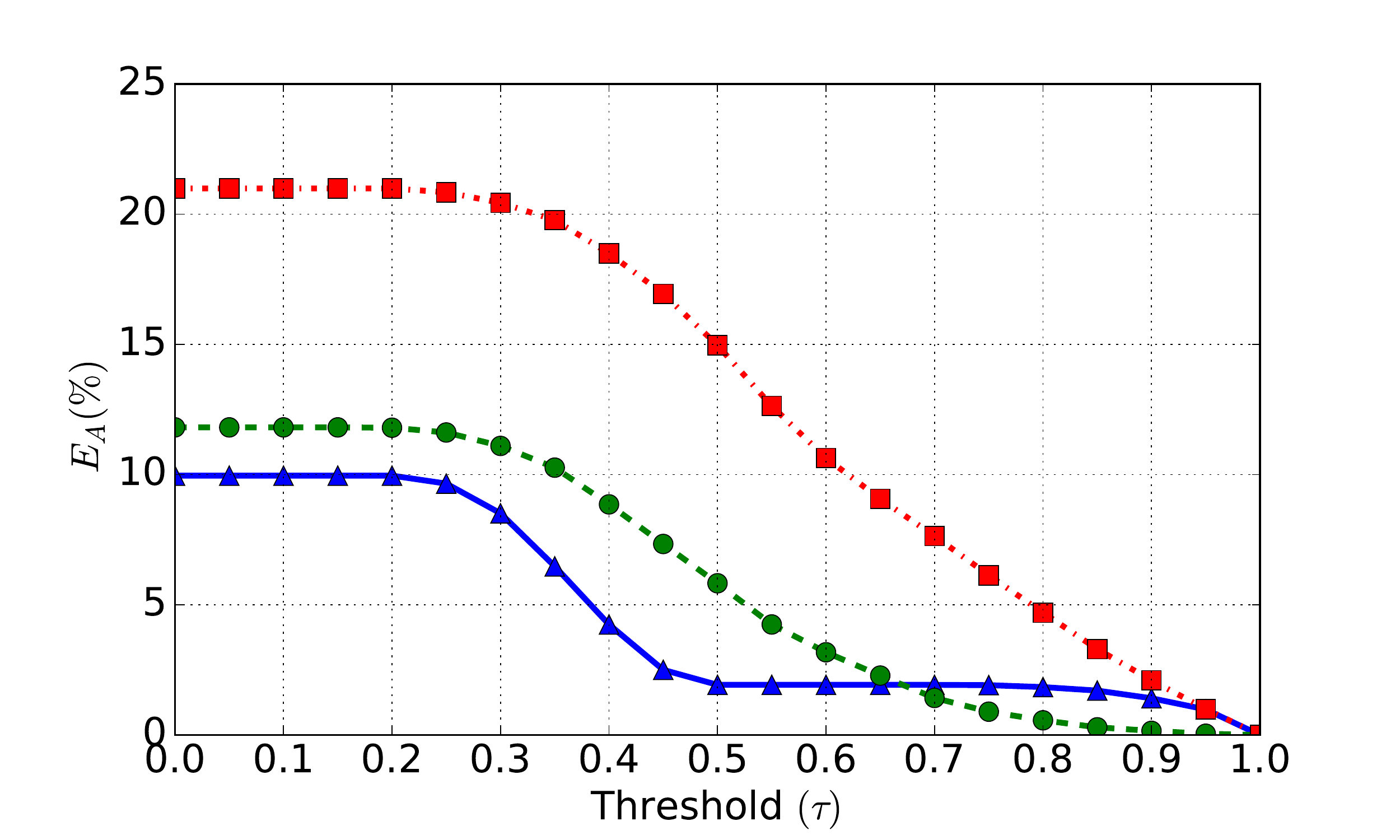}}
\caption{Error rates $E_{D}$ on clean test samples, and error rates $E_{A}$ on their corresponding adversaries, as a function of threshold ($\tau$), for the MNIST and CIFAR-10 datasets.}

\label{ErrExprmnt-woGAN}
\end{figure}

Fig.~\ref{ErrExprmnt-woGAN} presents error rates $E_D$ (Eq.~\ref{ED}) and $E_A$ (Eq.~\ref{EA}) on the MNIST and CIFAR-10 datasets with clean samples and three types of adversaries. For naive CNN* and the pure ensemble, error rates of different types of adversaries ($E_{A}$) decrease monotonically as the threshold increases. However, the error rates of adversaries by specialists+1 ensemble are not monotonically decreased. As confidences for most of the misclassified adversaries by specialists+1 ensemble is lower than $0.5$, rejection of low confidence predictions at this threshold results in a significant reduction of adversaries error $E_{A}$ in comparison to naive CNN* and the pure ensemble using this threshold. Accordingly, it can be confirmed that specialists+1 can shift most of the misclassified adversaries to low confidence, thus they are being rejected. 

Some clean samples that can be correctly classified with high confidence by a CNN are rejected by the specialists+1 ensemble due to their low confidence, thus increasing slightly $E_{D}$ at a threshold $0.5$. Note that increasing the threshold to a higher value causes rejection of a vast majority of adversaries as well as more clean samples. This thus requires a trade-off between keeping rejection rate of clean test samples low vs rejecting adversaries that would otherwise fool the networks.

\begin{figure}
\centering
\subfigure[Confidence densities for naive CNN*]{\label{MNIST:Naive}\includegraphics[width=0.49\textwidth, height=4cm]{MNIST_Naive}}\hfil
\subfigure[Rejection rate as a function of threshold for naive CNN*]{\label{MNIST:RNaive}
\includegraphics[width=0.49\textwidth]{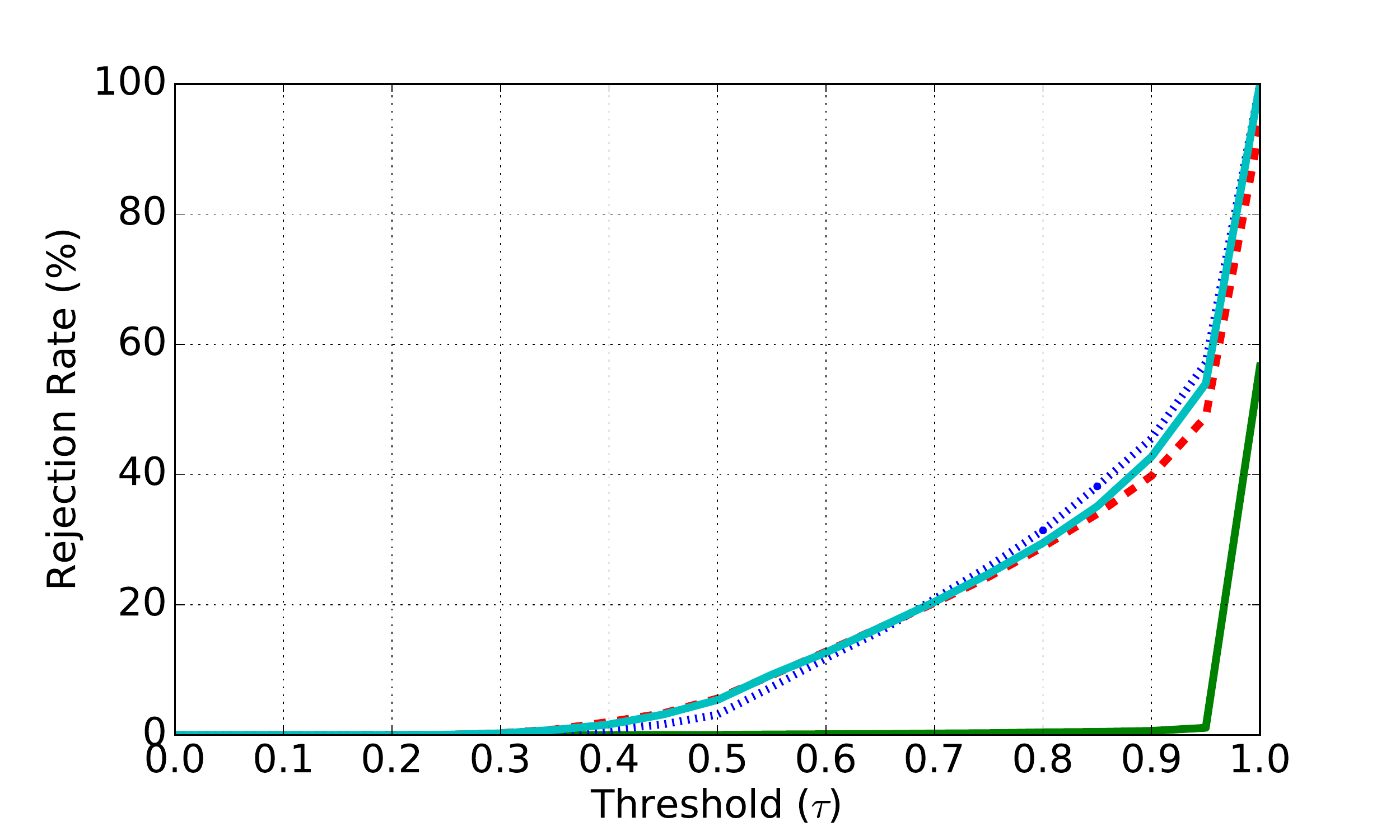}}
\subfigure[Confidence densities for pure ensemble]{\label{MNIST:pure}
\includegraphics[width=0.49\textwidth]{MNIST_Pure}}\hfil
\subfigure[Rejection rate as a function of threshold for pure ensemble]{\label{MNIST:Rpure}
\includegraphics[width=0.49\textwidth]{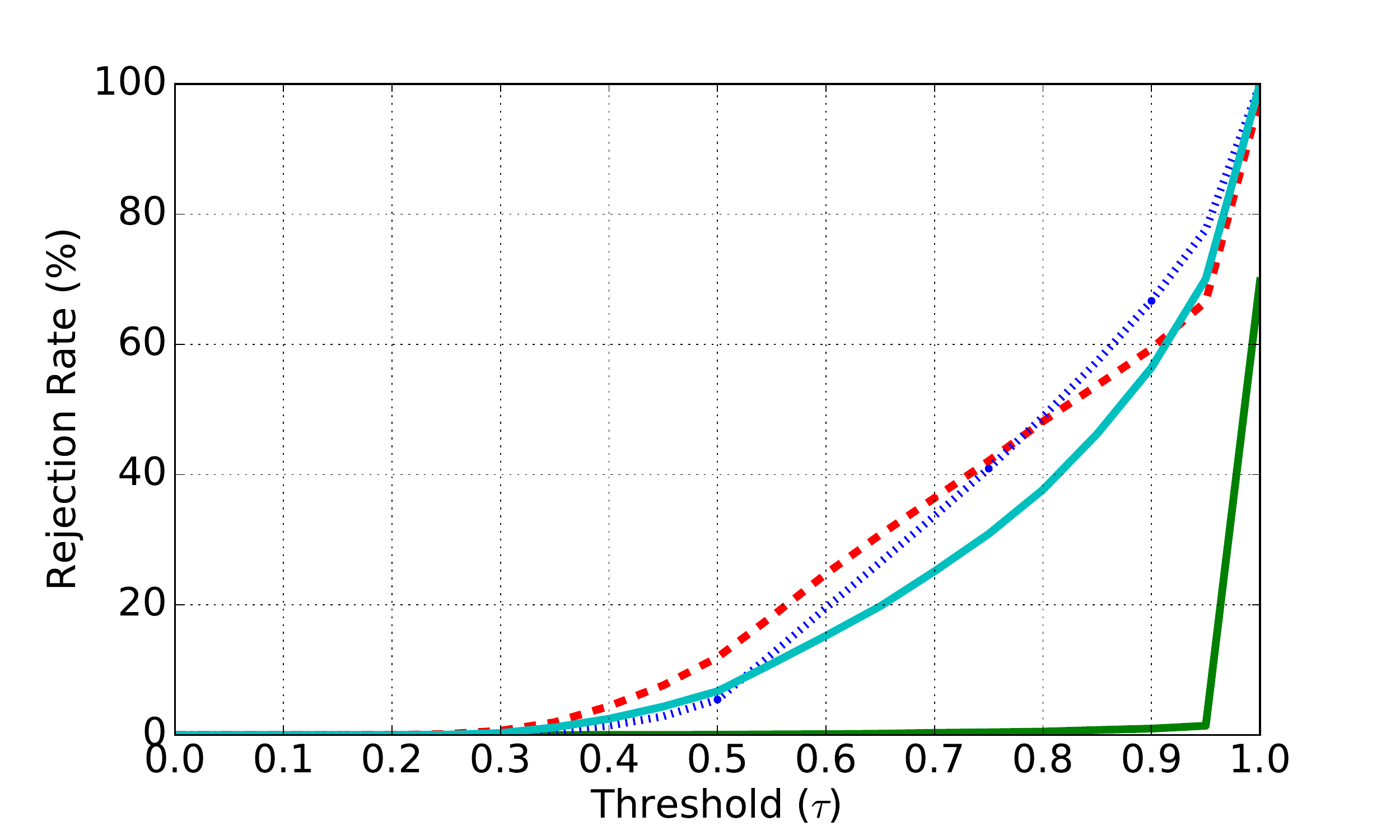}}
\subfigure[Confidence densities for specialists+1]{
\label{MNIST:Our}
\includegraphics[width=0.49\textwidth]{MNIST_Our}}\hfil
\subfigure[Rejection rate as a function of threshold for specialists+1]{\label{MNIST:Rour}
\includegraphics[width=0.49\textwidth]{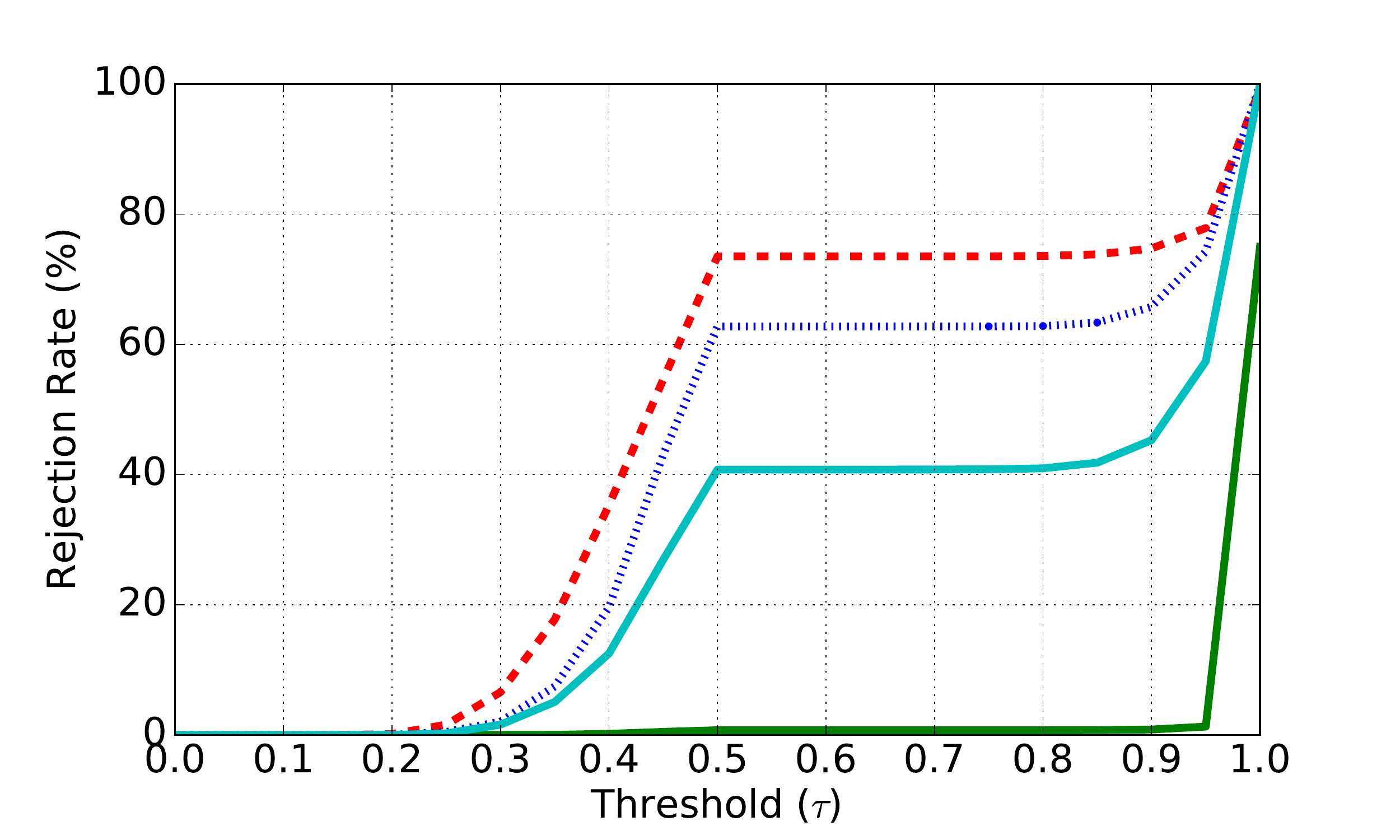}}
\caption{Confidence densities and rejection rates as a function of threshold on confidence for MNIST clean test samples and their adversaries depicted for different approaches.}
\label{DenMNIST}
\end{figure}

\begin{figure}
\centering
\subfigure[Confidence densities for naive CNN*]{
\includegraphics[width=0.49\textwidth]{cifar_Naive_wo}}\hfil
\subfigure[Rejection rate as a function of threshold for naive CNN*]{
\includegraphics[width=0.49\textwidth]{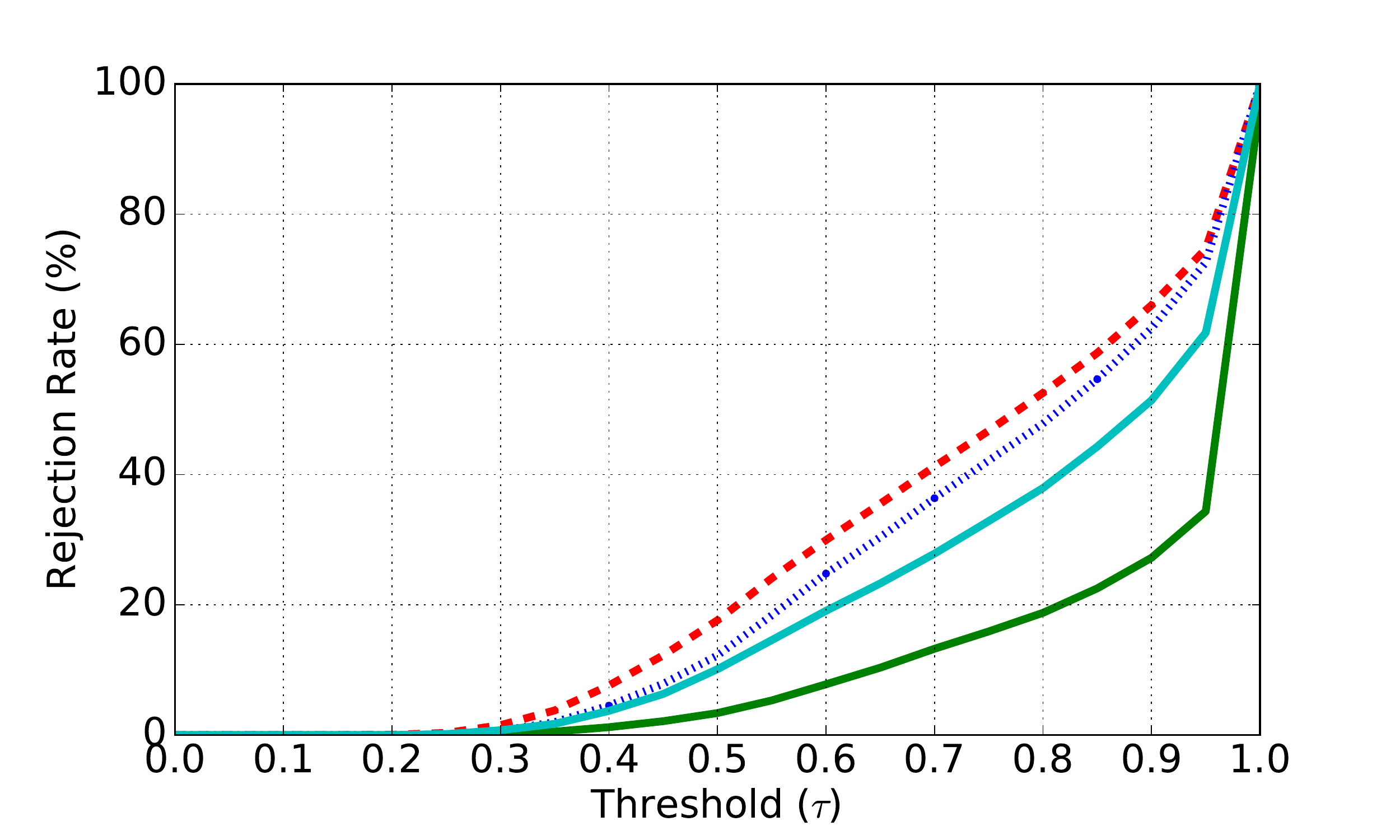}}
\subfigure[Confidence densities for pure ensemble]{
\includegraphics[width=0.49\textwidth]{cifar_Pure_wo}}\hfil
\subfigure[Rejection rate as a function of threshold for pure ensemble]{
\includegraphics[width=0.49\textwidth]{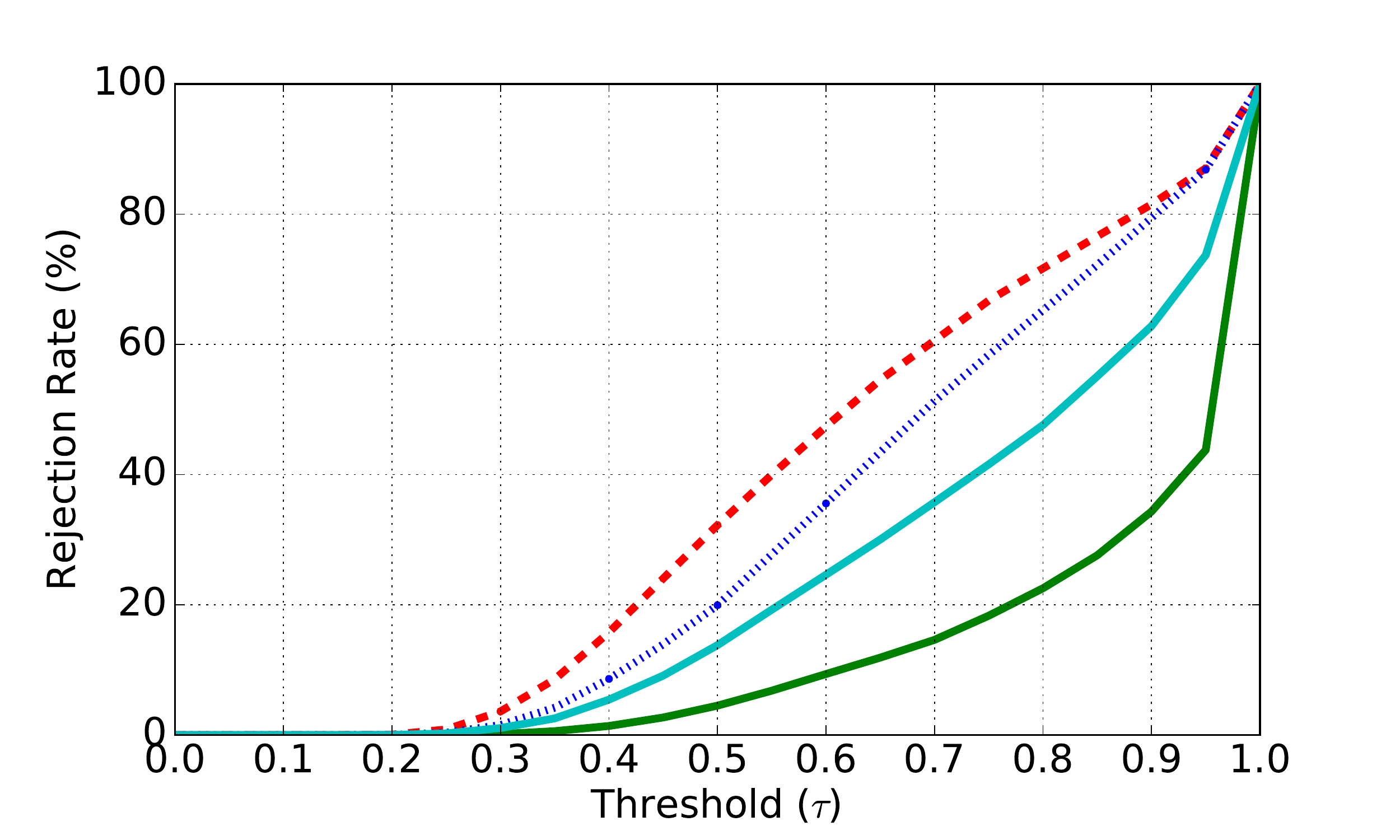}}

\subfigure[Confidence densities for specialists+1]{\label{Cifar-Our}
\includegraphics[width=0.49\textwidth]{cifar_Our_wo}}\hfil
\subfigure[Rejection rate as a function of threshold for specialists+1]{\label{Cifar-ROur}
\includegraphics[width=0.49\textwidth]{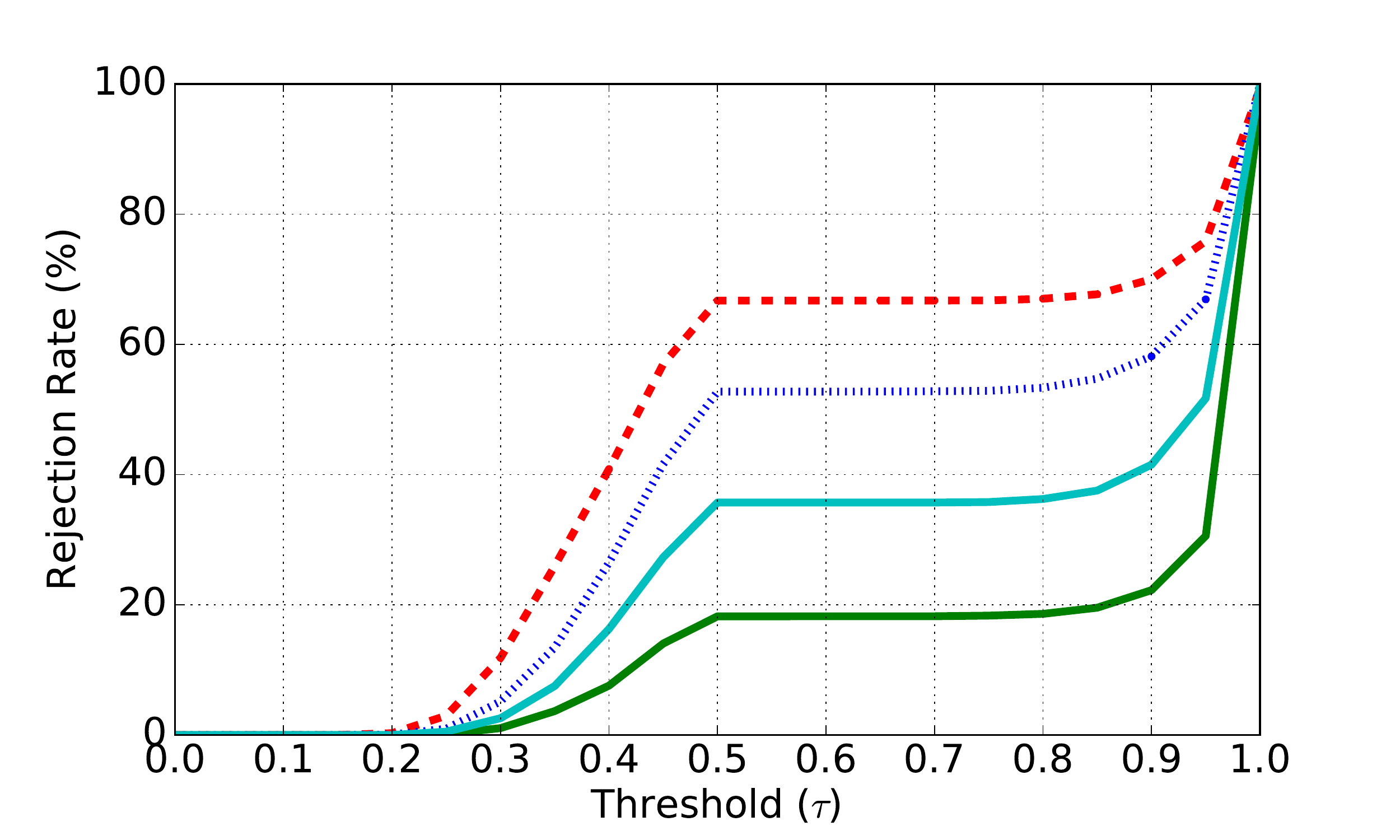}}
\caption{Confidence densities and rejection rates as a function of threshold on confidence for CIFAR-10 clean test samples and their adversaries depicted for different approaches.}
\label{DenCifar}
\end{figure}

Moreover, from the error rates of FGS and DF adversaries ($E_{A}$) at threshold zero (Fig.~\ref{ErrExprmnt-woGAN}), it can be seen that FGS and DF adversaries can severely fool the new models since they are transferable, i.e. their cross-model generalization property, while adversaries by~\citet{szegedy2013intriguing} are less generalized across different models. So, a remarkable number of Szegedy adversaries can be correctly and confidently classified by the models that are different from GA-CNN (the adversaries generative model). Notice that $E_{A}$ of Szegedy adversaries by specialists+1 does not change considerably after threshold $0.5$. Since most of the high confidence predictions for this type of adversaries (shown by the cyan pick at the high confidence in Fig.~\ref{MNIST:Our} and Fig.~\ref{Cifar-Our}) are correctly and confidently classified by specialists+1 ensemble.

%\begin{figure}
%\centering
%\subfigure[{$E_{D}(\%)$ on MNIST clean test set}]{\label{Err_MNISTcln}\includegraphics[width=0.49\textwidth]{MNIST_Er_test_new2}}
%\hfil
%\subfigure[{$E_{A}(\%)$ on MNIST FGS adversaries}]{\includegraphics[width=0.49\textwidth]{MNIST_Er_A_FGS2}}\\
%\subfigure[{$E_{A}(\%)$ on MNIST DF adversaries}] {\includegraphics[width = 0.49\textwidth]{MNIST_Er_A_DF2}}\hfil
%\subfigure[{$E_{A}(\%)$ on MNIST adversaries by~\citet{szegedy2013intriguing}}]{\includegraphics[width = 0.49\textwidth]{MNIST_Er_A_LBFGS2}}
%\\
%
%\subfigure[{$E_{D}(\%)$ on CIFAR-10 clean samples}]{\label{Err_Cifarcln}\includegraphics[width=0.49\textwidth]{CIFAR10_Er_test_new2}}
%\hfil
%\subfigure[{$E_{A}(\%)$ on CIFAR-10 FGS adversaries}]{\includegraphics[width = 0.49\textwidth]{CIFAR10_Er_A_FGS2}}\\
%\subfigure[{$E_{A}(\%)$ on CIFAR-10 DF adversaries}]{\includegraphics[width = 0.49\textwidth]{CIFAR10_Er_A_DF2}}\hfil
%\subfigure[{$E_{A}(\%)$ on CIFAR-10 adversaries by~\citet{szegedy2013intriguing}}]{\includegraphics[width = 0.49\textwidth]{CIFAR10_Er_A_LBFGS2}}
%\caption{Error rates $E_{D}$ on clean test samples, and error rates $E_{A}$ on their corresponding adversaries, as a function of threshold ($\tau$), for the MNIST and CIFAR-10 datasets.}\label{ErrExprmnt}
%\end{figure}

%=========== Without GAN ===========

For a better insight, the rejection rate as a function of the threshold on confidence for MNIST and CIFAR-10 are shown in association with their distributions of confidence in Fig.~\ref{DenMNIST} and \ref{DenCifar}, respectively. According to these observations, specialists+1 successfully provides significantly lower confidence for most of the adversaries, regardless of their types, than naive CNN* and the pure ensemble. Therefore, its rejection rate curves of adversaries are increasing at lower confidence and reach to some picks very fast at a threshold of $0.5$. However, the rejection rates of adversaries by the pure ensemble are monotonically increasing by increasing the threshold, and reach to their picks at a higher threshold, when a mix of both clean and adversaries samples are being rejected. Also, as it can be seen from Fig.~\ref{MNIST:RNaive}, \ref{MNIST:Rpure}, and \ref{MNIST:Rour} that rejection rate curves for MNIST correctly classified clean test samples by these three frameworks are mostly similar. 

\end{document}